\pdfoutput=1

\documentclass[11pt]{article}

\usepackage[preprint]{acl}

\usepackage{times}
\usepackage{latexsym}

\usepackage[T1]{fontenc}

\usepackage[utf8]{inputenc}

\usepackage{microtype}

\usepackage{inconsolata}

\usepackage{graphicx}

\usepackage{tabularx}
\usepackage{booktabs}
\usepackage{float}
\usepackage{amsmath, amsfonts}
\usepackage{placeins}  %
\usepackage{adjustbox}
\usepackage[normalem]{ulem}
\useunder{\uline}{\ul}{}

\title{OpenNER 1.0: Standardized Open-Access Named Entity Recognition Datasets in 50+ Languages}

\author{
  Chester Palen-Michel \and Maxwell Pickering \and Maya Kruse \\
  \and \textbf{Jonne Sälevä} \and \textbf{Constantine Lignos} \\
  Michtom School of Computer Science \\
  Brandeis University \\
  \texttt{\{cpalenmichel,pickering,mayakruse,jonnesaleva,lignos\}@brandeis.edu} \\
}

\begin{document}
\maketitle
\begin{abstract}
We present OpenNER 1.0, a standardized collection of openly-available named entity recognition (NER) datasets.
OpenNER contains 36 NER corpora that span 52 languages, human-annotated in varying named entity ontologies.
We correct annotation format issues, standardize the original datasets into a uniform representation with consistent entity type names across corpora, and provide the collection in a structure that enables research in multilingual and multi-ontology NER.
We provide baseline results using three pretrained multilingual language models and two large language models to compare the performance of recent models and facilitate future research in NER.
We find that no single model is best in all languages and that significant work remains to obtain high performance from LLMs on the NER task.
OpenNER is released at \url{https://github.com/bltlab/open-ner}.\footnote{The archival version of this paper is located at:\\ \url{https://aclanthology.org/2025.emnlp-main.1708/}}
\end{abstract}

\section{Introduction}

\begin{figure}[t!]
    \includegraphics[width=\columnwidth]{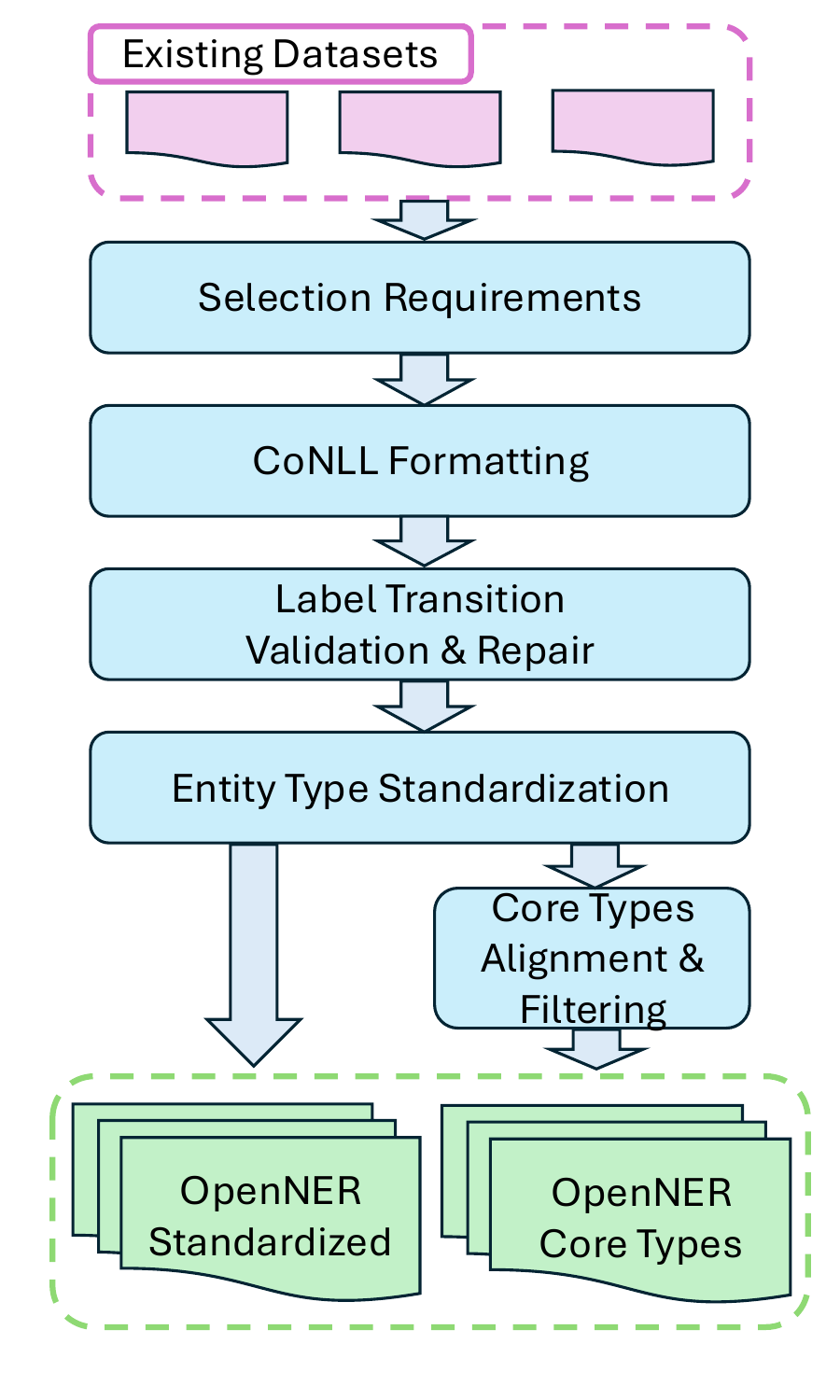}
    \caption{The processing pipeline for OpenNER. Existing datasets (magenta) pass through a series of stages of standardization (blue) to produce two final versions of the dataset (green).}    
    \label{fig:openner-flowchart}
\end{figure}

In the 25+ years following the 7th Message Understanding Conference \citep[MUC-7,][]{chinchor-1998-overview}, there has been steady development of new datasets for the task of named entity recognition (NER).
While the CoNLL 2002--3 shared task datasets \citep{tjong-kim-sang-2002-introduction,tjong-kim-sang-de-meulder-2003-introduction} and OntoNotes \citep{hovy-etal-2006-ontonotes} are perhaps the most famous, dozens of corpora have been released since in many languages.

Despite the constant release of new datasets, there is no straightforward way for researchers to work with multiple NER corpora.
There is no central repository of NER data, and many of the datasets appearing on lists of NER resources are not readily usable.
Many datasets are not consistently formatted and use a variety of chunk encodings (IOB, BIO, etc.), often without documentation.

This paper presents OpenNER 1.0, a first-of-its-kind multilingual, multi-ontology collection of openly-available human-annotated NER datasets to enable painless research into NER beyond the scale of a single corpus.
OpenNER is the largest collection of human-annotated NER data created to-date.
As new datasets are created and released, we intend for OpenNER to grow in the number of languages covered.

The process of creating OpenNER is shown in Figure~\ref{fig:openner-flowchart}.
We release OpenNER in two versions.
The \emph{standardized} version contains all datasets in their original named entity ontologies, with the entity type names mapped to a standard set (e.g. \texttt{PER} is used for ``person'' in all datasets).
The \emph{core types} version contains all datasets but only includes person, location, and organization entity types.

OpenNER is released at \url{https://github.com/bltlab/open-ner}.
The repository contains all code needed to assemble and preprocess all datasets.
The repository README contains links to where any other copies of the data are hosted (Hugging Face, etc.).
The OpenNER collection is licensed under the Creative Commons CC BY license; however, all datasets contained in it are licensed under their own licensing terms, some of which prohibit commercial usage.

\section{Data Sources}

\begin{table*}[t]
\centering
\footnotesize
\adjustbox{max width=\textwidth, max height=\textheight}{
\begin{tabular}{@{}llllll@{}}
\toprule
Corpus         & Ontology  & Source                                  & Corpus                      & Ontology  & Source                                  \\ \midrule
AnCora        & CoNLL     & \citet{taule-etal-2008-ancora}            & L3Cube MahaNER       & CoNLL     & \citet{litake-etal-2022-l3cube}           \\
AQMAR         & CoNLL     & \citet{mohit-etal-2012-recall}            & MasakhaNER           & CoNLL     & \citet{adelani-etal-2021-masakhaner}      \\
ArmanPersoNER & CoNLL     & \citet{poostchi-etal-2016-personer}       & MasakhaNER 2         & CoNLL     & \citet{adelani-etal-2022-masakhaner}      \\
BarNER        & CoNLL     & \citet{peng-etal-2024-sebastian}          & NEMO2                & CoNLL     & \citet{bareket-tsarfaty-2021-neural}      \\
CONLL-02      & CoNLL     & \citet{tjong-kim-sang-2002-introduction}  & NorNE                & CoNLL     & \citet{jorgensen-etal-2020-norne}         \\
DaNE          & CoNLL     & \citet{hvingelby-etal-2020-dane}          & RONEC                & OntoNotes & \citet{dumitrescu-avram-2020-introducing} \\
EIEC          & CoNLL     & \citet{alegria2006lessons}                & SLI Galician Corpora & CoNLL     & \citet{agerri-etal-2018-developing}       \\
elNER         & OntoNotes & \citet{elner-2020}                        & ssj500k              & CoNLL     & \citet{dobrovoljc-etal-2017-universal}    \\
EverestNER    & CoNLL     & \citet{niraula2022named}                  & ThaiNNER             & OntoNotes & \citet{buaphet-etal-2022-thai}            \\
GermEval      & CoNLL     & \citet{benikova2014germeval}              & TurkNLP              & CoNLL     & \citet{luoma-etal-2020-broad}             \\
HiNER         & CoNLL     & \citet{murthy-etal-2022-hiner}            & Tweebank             & CoNLL     & \citet{jiang-etal-2022-annotating}        \\
hr500k        & CoNLL     & \citet{ljubesic-etal-2016-new}            & UNER                 & CoNLL     & \citet{mayhew-etal-2024-universal}        \\
Japanese-GSD  & OntoNotes & \citet{asahara-etal-2018-universal}       & WikiGoldSK           & CoNLL     & \citet{suba-etal-2023-wikigoldsk}         \\
KazNERD       & OntoNotes & \citet{yeshpanov-etal-2022-kaznerd}       & WNUT17               & CoNLL     & \citet{derczynski-etal-2017-results}      \\
KIND          & CoNLL     & \citet{paccosi-palmero-aprosio-2022-kind} &                      &           &                                           \\ \bottomrule
\end{tabular}
}
\caption{Dataset sources and whether the entity type set is more similar to the CoNLL or OntoNotes ontologies.}
\label{tab:sources}
\end{table*}

\subsection{Selection Requirements}
\label{sec:reqs}
The requirements we set for inclusion of corpora in OpenNER are as follows. 

\paragraph{Openly-Accessible}
First, all datasets must be truly openly-accessible such that they can be easily and legally accessed on the open internet, without requiring the user to request the data or sign an agreement.
We do not include datasets that are ``available by request'' because our goal is to create a benchmark dataset that anyone can automatically run.\footnote{We have also found that many datasets that are only available by request have been collected in ways that potentially violate the terms of use or copyright of data sources.}
While all datasets we include are publicly available, some do restrict commercial usage.

\paragraph{Human Annotation} 
Second, the data must have been manually-annotated using explicit annotation guidelines; we do not include any ``silver-standard'' datasets where all or part of the annotation was automatically generated \citep[e.g.][]{fetahu-etal-2023-semeval,pan-etal-2017-cross,zhou2023universalner}.

\paragraph{General Purpose Ontology}
For the initial release of OpenNER, the annotation must center around traditional named entities, such as persons, locations, organizations, works of art, etc.
While we acknowledge their importance, we did not include corpora for chunk extraction in specific domains such as biomedical data or legal cases.
Adding these domains presents additional challenges for entity type standardization since they are less likely to have overlap with more generic NER entity types.
We leave the incorporation of such datasets to future work as it requires significant additional research.

We do not require any specific entity types to be included in the datasets; we include all types annotated in the original datasets, although we do rename some types to standardize them across corpora, for example renaming all variants of the person type (e.g., \texttt{PERSON}, \texttt{PERS}, \texttt{PER}) to \texttt{PER}.
We take a different approach than Universal NER (UNER) \citep{mayhew-etal-2024-universal} in that our goal is to include as many existing datasets as possible, despite their annotation differences, rather than producing new datasets with uniform annotation.

\paragraph{Sufficient Data}
We require that there be enough data to create training and test datasets to support experiments.
This excludes some small test-only corpora, such as the Europarl annotations \citep{agerri-etal-2018-building}, which are significantly smaller than most of the other included datasets. 
Similarly, UNER \citep{mayhew-etal-2024-universal} contains a number of test-only datasets that we did not include. 

\paragraph{Tokenization and Formatting}
Finally, the data must be available in a tokenized format; if not already ``CoNLL-style,'' one that can be straightforwardly converted into it.
We tried to accept as many corpora as possible, correcting a substantial number of formatting and entity encoding errors.
While we are interested in including datasets that do not provide tokenization, doing so would require either performing word segmentation for every corpus and aligning it to the annotation---an error-prone and lossy process---or a new set of tools for preprocessing and training NER models, as most models take pretokenized data as input.

\subsection{Datasets Included}

We include 36 corpora spanning 52 languages in OpenNER.
Most of the datasets use a variant of the CoNLL-02 ontology \citep{tjong-kim-sang-2002-introduction}, and a few are derived from OntoNotes \citep{hovy-etal-2006-ontonotes} or develop customized ontologies.
As seen in Table~\ref{tab:stats}, the datasets span a range of language families and differing numbers of entity types. 
We categorize the corpora as following either a CoNLL- or OntoNotes-derived ontology in Table \ref{tab:sources}, which provides names and citations for all included datasets.
The CoNLL-02 corpus \citep{tjong-kim-sang-2002-introduction} consists of Spanish and Dutch newswire data and introduces the \texttt{LOC}/\texttt{ORG}/\texttt{PER}/\texttt{MISC} tagset adapted by many other corpora in this collection.
The majority of corpora in OpenNER follow a type ontology similar to that of CoNLL-02 with \texttt{PERSON}, \texttt{LOCATION}, \texttt{ORGANIZATION}, and \texttt{MISC}. 

Some CoNLL-inspired corpora leave out \texttt{MISC} \citep[e.g.][]{mayhew-etal-2024-universal}, while others replace \texttt{MISC} with other types. ArmanPersoNER \citep{poostchi-etal-2016-personer} adds \texttt{EVENT}, \texttt{FACILITY}, and \texttt{PRODUCT}, while MasakhaNER \citet{adelani-etal-2021-masakhaner} adds \texttt{DATE}. 
Other corpora follow the OntoNotes ontology but collapse types \citep[RONEC,][]{dumitrescu-avram-2020-introducing}, add types \citep[Japanese-GSD,][]{asahara-etal-2018-universal}, or use a subset \citep[ThaiNNER,][]{buaphet-etal-2022-thai}.
More detail about the included datasets, including variations in entity types, are provided in Appendix~\ref{app:ontologies}.

\subsection{Datasets not Included}

Unfortunately, some datasets could not be included in our collection for a variety of reasons.
Many datasets require the user to request either the annotations or the text.
The CoNLL-03 shared task \citep{tjong-kim-sang-de-meulder-2003-introduction} and OntoNotes \citep{hovy-etal-2006-ontonotes} datasets use text that cannot be freely distributed; legal use of the data requires that the source text be requested from NIST and the LDC respectively.
The data for the EVALITA 2009 Italian NER shared task \citep{speranza2009named} was only available by request.
The Wojood Arabic NER dataset \citep{jarrar-etal-2022-wojood} only has a sample of data publicly available a at the time this research was performed; the remainder of the dataset is only available upon request.
Datasets that require payment, such as the LORELEI language packs for less-resourced languages \citep{strassel-tracey-2016-lorelei}, also could not be included because they are not freely available.

We cannot easily convert datasets to CoNLL format without an authoritative tokenization of the data. 
This unfortunately excludes some datasets which are otherwise good candidates for inclusion. 
Datasets which report mentions as character offsets but without tokenization could not be included, such as the MEN corpus of Malaysian English news \citep{chanthran-etal-2024-malaysian} and the DANSK corpus of multi-domain Danish \citep{enevoldsen_dansk}. 
Similarly, the multilingual SlavicNER corpus reports a list of mentions with character offsets for each source document, but without tokenization \citep{piskorski-etal-2024-cross}. 
The ENP-NER corpus of historical Chinese newspapers reports character-level tags \citep{blouin-etal-2024-dataset}.

We did not include corpora for specialized domains such as biomedical data \citep{byun-etal-2024-korean}, paper abstracts \citep{phi-etal-2024-polynere,alkan-etal-2024-enriching}, and industrial documents \citep{li-etal-2024-corpus}.

We only include datasets created using human annotation.
Although WikiAnn \citep{pan-etal-2017-cross} is often used as a multilingual NER benchmark, it is a ``silver-standard'' dataset and uses automatically-created labels.
We did not include MultiCoNER \citep{malmasi-etal-2022-semeval} as it has not been hand-annotated, but rather extracted from text that is linked to articles corresponding to entity types. We do not include NerKor+Cars-OntoNotes++ \citep{novak-novak-2022-nerkor} because it uses a semi-automatic labeling approach where not all labels are manually checked.
As there are many popular and widely-known NER datasets that are not human-annotated, our goal with NER was to provide a complementary dataset that only contains human annotation.

Some candidate datasets were not included because of pervasive dataset quality issues or annotation errors that were too onerous to repair.
Details for these datasets are included in Appendix \ref{app:not-included-data}.
Finally, we could not include the datasets for many older papers because they were no longer available.

\section{Standardization}

\subsection{CoNLL Formatting}
We require all included datasets to be converted to the CoNLL format with BIO mention encoding and UTF-8 text encoding.
The CoNLL format represents labeled sequences with one token per line, with sentences separated by newlines.
The type label and any other metadata pertaining to the token appear on the same line as the token, separated by whitespace.
We had to modify the text encoding and file formats used by several corpora; details are provided in Appendix Section~\ref{sec:formatting-corrections}.

\subsection{Label Transition Validation}

We corrected label transition errors---failures to correctly follow the BIO, IOB, etc. encoding schemes---automatically when possible, and manually when required. 
Repairing invalid label sequences involved validating with SeqScore \citep{palen-michel-etal-2021-seqscore} and manually reviewing the validation errors. 
If the errors all appeared to be safely repaired with SeqScore's repair functionality, automated repairs were performed.
In some cases, manual repairs were conducted, and those repairs are detailed in Appendix Section~\ref{sec:repair}.

\subsection{Entity Type Standardization}
Once all datasets were correctly BIO-encoded, we standardized the entity types in order to have a consistent set of entity type labels across all datasets.

We adopted the following conventions for named entity types.
Each type is written as capitalized letters, with underscores used to separate words in multi-word names (e.g. \texttt{PET\_NAME}).
Any sub-tags are written with a hyphen, following the ``dash tag'' style (e.g. \texttt{LOC-DERIV}).
When there is a commonly-used short form for a type (e.g. \texttt{ORG}), we map longer versions of the name to it.
For example, there are six different ways that the ``organization'' entity type is named across different corpora: 
\texttt{ORG}, \texttt{Organization}, \texttt{ORGANIZATION}, \texttt{ORGANISATION}, \texttt{org}, \texttt{NEO}.
We standardize them all to \texttt{ORG}.
Similar but non-identical entity types, such as \texttt{DATETIME} and \texttt{TIME}, are left separate.

This standardization process preserves all annotation in the original datasets.
No name mentions are removed, and within each corpus, no types are combined.
This process creates the most uniform set of types possible across all of the datasets and allows for better usability; however, users should be aware that annotation guidelines still vary across corpora, leading to the span or entity type for similar mentions across two different corpora to differ. 
OpenNER provides the first easy way to explore these differences in annotation at scale. 
Appendix Table~\ref{tab:map-unify-types} gives the full mapping of types used in the standardization process.
There are 60 unique entity types across 2,816,304 total mentions.

\subsection{Core Types}

\begin{table}[tb]
\centering
\footnotesize
\begin{tabular}{ll}
\toprule
Before                                  & After  \\ 
 Mapping                                 & Mapping \\ \midrule
PER, PER-PART                                  & PER           \\ \midrule
LOC, GPE, GPE-LOC, LOC-PART,                   &            \\
GPE-ORG, FACILITY                               & LOC \\ \midrule
ORG, ORG-PART,                                  &            \\
 CORPORATION, GROUP                                  & ORG           \\\bottomrule
\end{tabular}
\caption{Entity types mapped to core types of PER, ORG, and LOC for the core types version of OpenNER.}
\label{tab:map-core-types}
\end{table}

We additionally provide a secondary version of OpenNER where we map entity types to a set of minimal core types---location (\texttt{LOC}), organization (\texttt{ORG}), and person (\texttt{PER})---and discard all other types.
This minimal ontology is useful for exploring commonalities across datasets and training multi-corpus and multilingual models.
Table~\ref{tab:map-core-types} gives the mapping of types used to create the core entity types version of the data, which was manually created after reviewing the annotation guidelines of each dataset.

While we have reduced all datasets to three common types, that does not mean that in every corpus those entity types are annotated in the same way.
For example, some corpora may ``tag by usage,'' such that if a sentence is about the physical location of a corporation it might be tagged as \texttt{LOC} instead of \texttt{ORG}, while others still use \texttt{ORG} in that instance.
Corpora may also differ in the extent of the annotated span.
OpenNER does not modify the original annotation beyond the mapping of types, so these differences across corpora persist in the core types version of the data.

\subsection{Dataset Statistics}

\begin{table}[tb]
\centering
\footnotesize
\begin{tabular}{lr}
\toprule
Entity Type & Count \\
\midrule
LOC & 464,426 \\
ORG & 247,745 \\
PER & 330,094 \\
\midrule
Total & 1,042,265 \\
\bottomrule
\end{tabular}
\caption{Counts of names of each entity type in the core types version of OpenNER.}
\label{tab:core_type_counts}
\end{table}

OpenNER includes annotations in 52 languages from a diverse set of language families that use 11 different scripts.
Table~\ref{tab:stats} gives the number of entity types and the number of training, validation, and test sentences for each language in each corpus.
Appendix Table~\ref{tab:langs} gives statistics and general information about the languages included in OpenNER.

Appendix Table~\ref{tab:entity_type_counts} gives the counts for every standardized entity type in our resource.
The most frequent type is \texttt{LOC}, with 412k mentions out of a total of 2.8M.
The rarest types are mostly dash tags (e.g. \texttt{EVENT-DERIV}) and a few very rarely-used types (e.g. \texttt{NON\_HUMAN}).
We include all entity types present in the original dataset, regardless of their frequency, but users of OpenNER may choose to exclude rare types in future evaluations.
Table~\ref{tab:core_type_counts} gives the entity types counts for the 1M mentions in the core types version of the data, where \texttt{LOC} remains the most frequent type.

\section{Experiments}

\begin{figure*}[tb]
    \includegraphics[width=\textwidth]{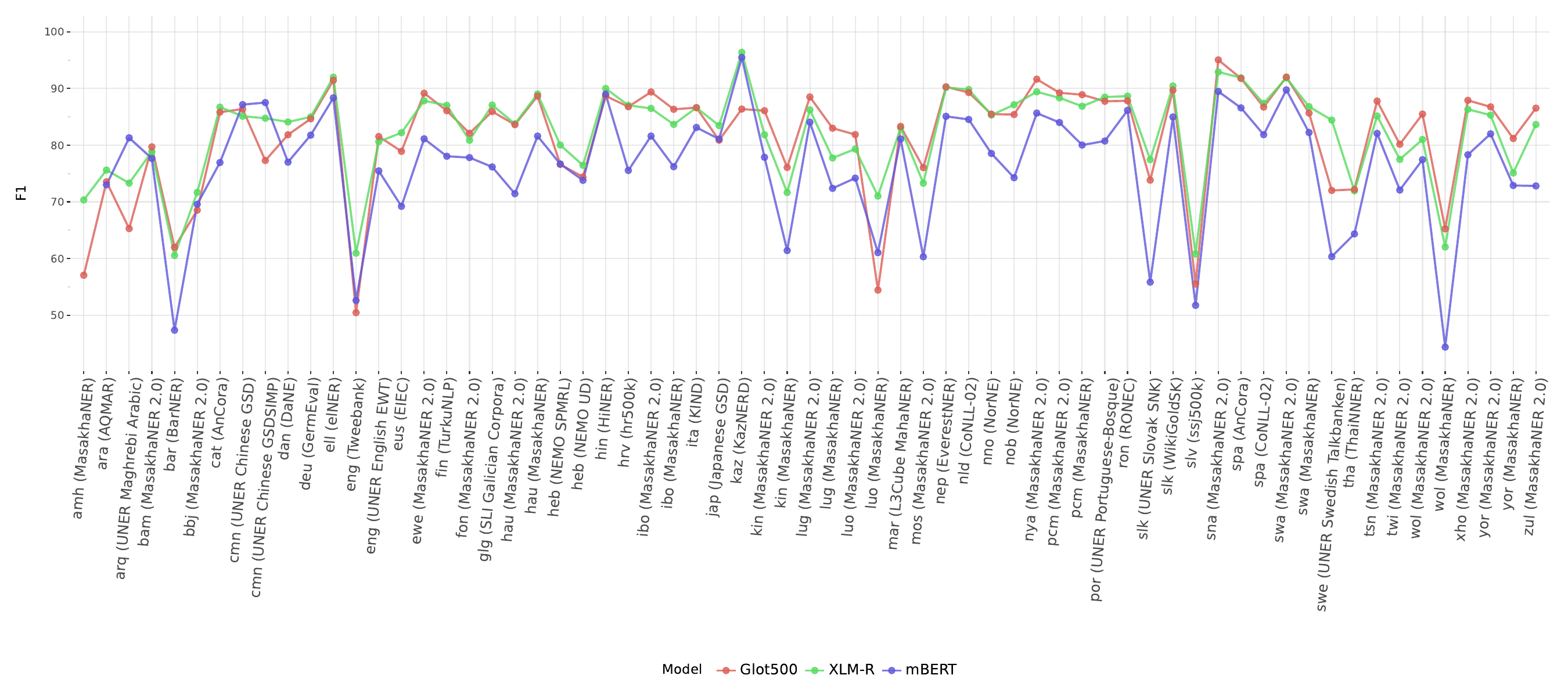}
    \caption{Mean F1 for each dataset-language combination, using all entity types present in each dataset. Models were fine-tuned individually on each dataset-language combination.}
    \label{fig:all-types-individual}
\end{figure*}

\begin{figure*}[tb]
    \includegraphics[width=\textwidth]{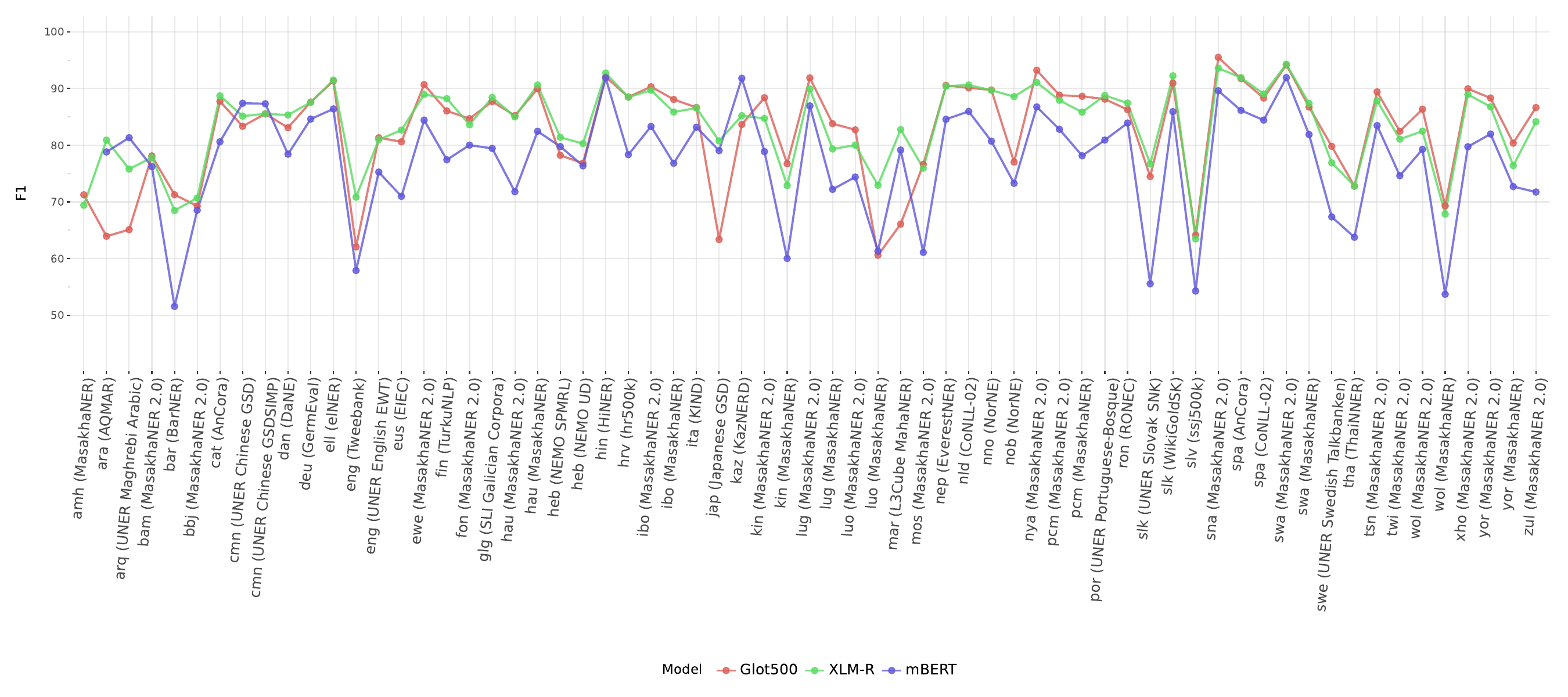}
    \caption{Mean F1 for each dataset-language combination, using only core entity types (location, organization, and person). Models were fine-tuned individually on each dataset-language combination.}    
    \label{fig:core-individual}
\end{figure*}

\begin{figure*}[tb]
    \includegraphics[width=\textwidth]{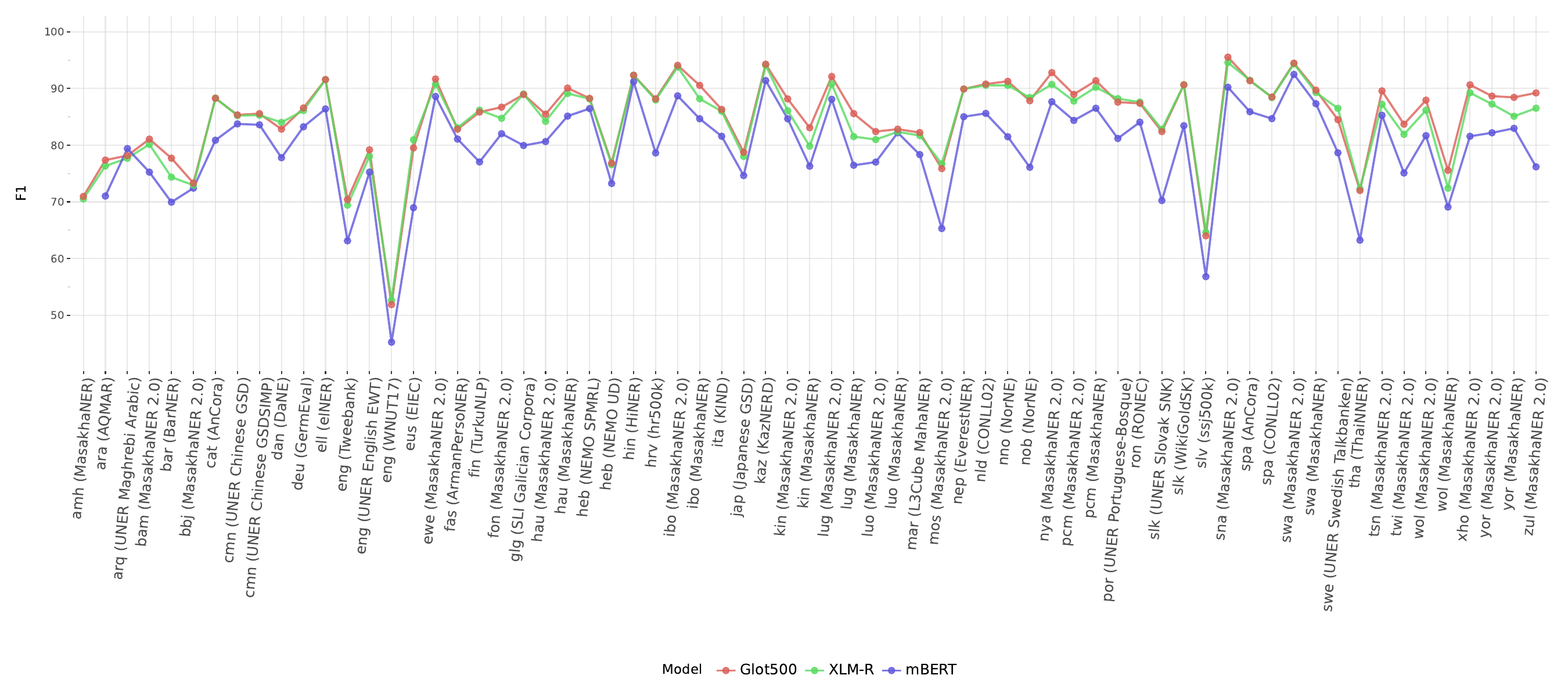}
    \caption{Mean F1 for each dataset-language combination, using only core entity types (location, organization, and person). Multilingual models were fine-tuned using all datasets and languages.}    
    \label{fig:core-multilingual}
\end{figure*}

In addition to providing a resource, we provide baselines using popular methods for performing NER in many languages.
We fine-tuned the XLM-RoBERTa Base \citep[XLM-R,][]{conneau-etal-2020-unsupervised}, mBERT \citep{devlin-etal-2019-bert}, and Glot500 \citep{imanigooghari-etal-2023-glot500} models.
We selected these models due to their popular usage in NER for less-resourced languages.\footnote{Due to a request from a reviewer, we performed a post-hoc evaluation of XLM-R Large. We found it performs worse than XLM-R Base despite being a larger model, although it is possible performance could be improved with further tuning.}
Many ``state of the art'' NER papers only evaluate on the CoNLL-03 English data or a few other high-resource language datasets; they also often rely on monolingual or few-language models, making adapting them to the 52 languages of OpenNER non-trivial.
We also benchmark two LLMs, Aya-Expanse \citep{dang2024ayaexpansecombiningresearch} and QwQ-32B-Preview \citep{qwq-32b-preview} using only the core types to explore LLM performance.

We experimented with three approaches to developing models.
The first involves training one model for each language in each dataset (67 language-dataset combinations in total), using the full set of entity types in each dataset.
The second involves training one model for each language in each dataset, but using only the core types (\texttt{LOC}, \texttt{ORG}, and \texttt{PER}).
The third involves training one multilingual model (or using in-context learning with an LLM) across all datasets and languages using only the core types.
For the multilingual model, we cap the training data at 32k sentences per language per dataset to mitigate data imbalances.

These experiments allow us to explore performance on both the original and core types ontologies, demonstrate the feasibility of multilingual NER models, and evaluate the performance of LLMs in a relatively simple NER ontology.

We report micro-averaged mention-level F1 for each model computed with SeqScore using the same method as the \texttt{conlleval} script.
We report the mean and standard error of the mean over 10 different training runs, each using a different random seed for initialization.
In total, we performed 4,050 fine-tuning runs; this large number is due to the sheer number of languages and datasets in OpenNER and the number of random seeds.
Hyperparameters and model details are discussed in Appendix Section \ref{app:hyperparams}.

\subsection{Individual Models on Full Ontologies}

Results for training individual models using the full ontology for each dataset are shown in Figure~\ref{fig:all-types-individual}, with all results in Appendix Table~\ref{tab:indiv-results}.
Averaged across all language-dataset combinations, the best performing model was XLM-R (F1 of 81.79), followed by Glot500 (80.96), and then mBERT (74.42).
While Glot500 performs best on many of the least-resourced languages, XLM-R performs better on average.
While mBERT generally performs worse than the other models, it scores substantially better in both Chinese datasets and does exceptionally well on the Maghrebi Arabic dataset, which is written in NArabizi, a method of writing Arabic in Latin script used in North Africa.
However, mBERT does not function at all in Amharic due to not being pretrained on the Ge'ez script.
We observe that there may be evidence of catastrophic forgetting in Glot500, as some higher-resourced languages like Spanish, Swedish, and English Tweebank underperform using Glot500 compared with XLM-R, which it is based on.

\subsection{Multilingual and Individual Models on Core Types}

\begin{figure}
    \centering
    \includegraphics[width=\linewidth]{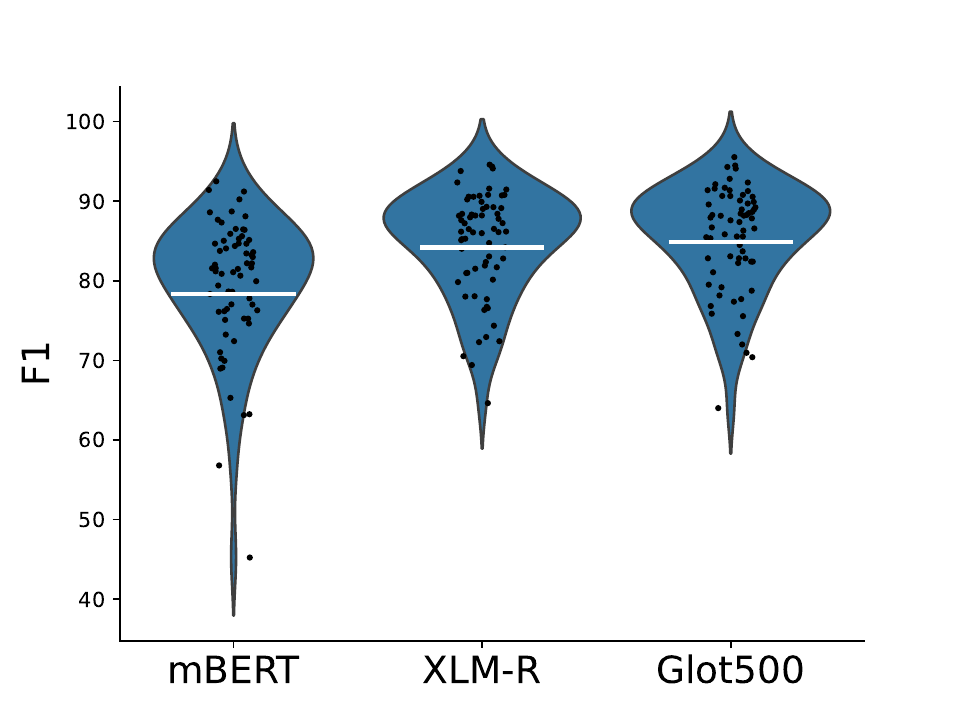}
    \caption{Violin plot of F1 distributions per model, with points depicting  mean F1 scores across random seeds for each language-dataset combination. White lines indicate means of all points per-model.}
    \label{fig:f1-violin-plot}
\end{figure}

Results for training individual and multilingual models using only the core types are shown in Figures~\ref{fig:core-individual} and \ref{fig:core-multilingual}, summarized in Table~\ref{tab:mean-across-corpora} and~Figure \ref{fig:f1-violin-plot}, with all results in Appendix Table~\ref{tab:multi-results}.
Glot500 again excels on the least-resourced languages and now also achieves the highest average performance.
The multilingual models often deliver better performance for less-resourced languages and cases where the exact same ontology is shared across datasets (e.g. MasakhaNER), while for many higher-resourced languages the best performance comes with individual-language models.
It is possible differences in annotation guidelines limit the multilingual model's performance in languages which do not have the same guidelines as others, while aiding transfer learning for datasets where guidelines are similar to others.

To assess the statistical significance of our main results, we compared each pair of the fine-tuned models using the non-parametric Wilcoxon signed-rank test with the core types models.
Paired comparisons were made for each of the three models using the 67 dataset-language combinations, comparing the mean F1 across random seeds for each model.
Each comparison is tested at the conventional 0.05 alpha level.
Since we are performing several hypothesis tests, we account for multiple comparisons using a Bonferroni correction and compare all \emph{p}-values to the corrected confidence threshold of $0.05/3$, equal to the alpha level of the test (0.05) divided by the number of tests (3).
This keeps the familywise false positive rate at 0.05.

When looking at the multilingual models (corresponding to the middle rows of Table~\ref{tab:mean-across-corpora}), on average, Glot500 outperforms XLM-R by 0.71 F1 ($p=1.64\cdot10^{-5} < 0.05/3$), Glot500 outperforms mBERT by 6.56 F1 ($p=1.34\cdot10^{-12} < 0.05/3$), and XLM-R outperforms mBERT by 5.85 F1 ($p=1.61\cdot10^{-12} < 0.05/3$).
These statistical tests confirm that the differences are statistically reliable across the language-dataset combinations.

When finetuning is conducted on individual languages rather than multilingually, the pattern changes slightly. On average, XLM-R now outperforms Glot500 by 0.44 F1 ($p=0.82 > 0.05/3$) though the result is not statistically significant.
Glot500 outperforms mBERT by 6.97 F1 ($p=1.96\cdot10^{-7} < 0.05/3$), which is similar to the previous difference of 6.56 F1.
XLM-R outperforms mBERT by 7.41 F1 ($p=4.21\cdot10^{-11} < 0.05/3$), which is larger than the 5.85 F1 difference observed earlier.

\subsection{LLMs}

\begin{table}[tb]
\centering
\small
\begin{tabular}{llr}
\toprule
Train Approach & Model       & Mean F1 \\
\midrule
Individual            & mBERT       & 75.69 \\
Individual            & XLM-R       & 83.10 \\
Individual            & Glot500     & 82.10 \\
\midrule
Multilingual        & mBERT       & 78.33 \\
Multilingual        & XLM-R       & 84.18 \\
Multilingual        & Glot500     & \textbf{84.89} \\
\midrule
In-context        & Aya-Expanse 32b & 60.55 \\ 
In-context        & QwQ 32b Preview         & 60.66 \\
\bottomrule
\end{tabular}

\caption{Mean F1 across each language-dataset combination for each approach, using only core types.}
\label{tab:mean-across-corpora}
\end{table}

We conducted baseline experiments with two LLMs, Aya-Expanse 32b and QwQ-32B-Preview using 5-shot demonstrations.
We chose these models because QwQ supports 29 languages and Aya-Expanse supports 23, and the models had similar numbers of parameters. 
The vLLM engine \citep{kwon2023efficient} was used for inference.
We evaluated using only the core types so that a standard prompt could be used across all datasets.

To score LLM output using traditional NER evaluation methods, one must map back to the original text, which poses many challenges.
Hallucinated tokens detected as names must either be discarded or penalized as false positives. 
There is also no guarantee that the generated labels will even be part of the ontology.
We conducted preliminary experiments with a handful of approaches for using LLMs to conduct NER, discussed further in Appendix \ref{app:llm}. 
Our preliminary experiments generally align with the findings of \citet{villena2024llmnerzerofewshotnamedentity}: few-shot demonstrations  perform better than zero-shot, and inline entity labeling generally outperforms JSON output.

For our LLM experiments, we used inline responses with prompts including 5 example demonstrations.
The prompt template, inspired by prompting techniques from \citet{wang-etal-2025-gpt}, is as follows:
\texttt{Find names of persons, organizations or locations. Label the following sentence with labels where the name is enclosed with the entity type PER, ORG or LOC and @@ \#\#. For example PER @@ John Smith \#\#, or ORG @@ Springfield University \#\# or LOC @@ United Kingdom \#\# .
Find named entities in the following sentence:
<<SENTTEXT>>}.

Full examples of the prompt with demonstrations are included in Appendix~\ref{app:llm}.
We conducted runs with 3 random seeds which were used for selecting the 5-shot demonstrations.
The demonstrations were the same for all sentences for each dataset-language combination.

The results, given in Table \ref {tab:mean-across-corpora} and Appendix Table~\ref{tab:llm-results}, show that LLM performance is substantially worse than the other methods we evaluate, consistent with other work showing performance lags behind encoder-only models with a classification head \citep{villena2024llmnerzerofewshotnamedentity,wang-etal-2025-gpt,xie-etal-2023-empirical}.
Curiously, the two models often scored identically---even though their predictions were not identical---suggesting that the in-context examples may be the limiting factor.

While we only explore comparatively simple prompting experiments in this work to establish LLM baselines, further methods with LLMs such as synthetic data generation \citep{santoso-etal-2024-pushing} and few-shot demonstration retrieval \citep{wang-etal-2025-gpt} are promising lines of research which OpenNER could facilitate. 
Whether and how to best leverage LLMs for NER, and in particular multilingual NER, remains an open problem, which we leave for future work.

\subsection{Discussion}

Overall, the results show that mBERT tends to perform worse than XLM-R or Glot500, but there are still cases where mBERT outperforms other models despite its age.
XLM-R and Glot500 perform similarly, with the former performing better when trained on individual languages, and the latter performing better when trained multilingually.
There does not currently appear to be a single best, one-size-fits-all model for these datasets.
While LLMs may eventually outperform sequence-labeling methods, further research is required to improve their performance.

\section{Future Work and Conclusion}

We believe OpenNER will facilitate future research in multilingual NER by drastically reducing the barrier to entry for researchers working with multiple NER datasets.
We have shown the potential for future experimentation with transfer learning and that there are challenges for training NER models that can handle multiple languages.

While OpenNER does not cover as many languages as ``silver standard'' (automatically annotated) datasets, it provides high-quality data in a smaller set of languages, many of them less-resourced.
We welcome the inclusion of additional datasets that we may have missed along with new datasets when they are created and released publicly.
We plan to release regular updates to include additional datasets as they are released and update our benchmarks with new methods.

\begin{table*}[t!]
\footnotesize
\centering
\vspace{-0.2in}
\adjustbox{max height=.50\textheight}{
\begin{tabular}{lllrrrrr}
\toprule
 &  &  &  & \multicolumn{4}{c}{Sentences} \\
\cmidrule(lr){5-8}
Corpus & Language & Code & Types & Train & Dev & Test & Total \\
\midrule
AnCora & Catalan & cat & 4 & 10,629 & 1,428 & 1,527 & 13,584 \\
AnCora & Spanish & spa & 6 & 11,374 & 2,992 & 2,983 & 17,349 \\
ArmanPersoNER & Persian & fas & 6 &  5,121 & 0 & 2,560 & 7,681 \\
AQMAR & Arabic & ara & 4 & 1,328 & 710 & 605 & 2,643 \\
BarNER & Bavarian German & bar & 23 & 2,869 & 338 & 370 & 3,577 \\
CoNLL-02 & Dutch & nld & 4 & 15,806 & 2,895 & 5,195 & 23,896 \\
CoNLL-02 & Spanish & spa & 4 & 8,323 & 1,915 & 1,517 & 11,755 \\
DaNE & Danish & dan & 4 & 4,383 & 564 & 565 & 5,512 \\
EIEC & Basque & eus & 4 & 2,552 & 0 & 842 & 3,394 \\
elNER & Greek & ell & 18 & 17,132 & 1,904 & 2,116 & 21,152 \\
EverestNER & Nepali & nep & 5 & 13,848 & 0 & 1,950 & 15,798 \\
GermEval & German & deu & 12 & 24,000 & 2,200 & 5,100 & 31,300 \\
HiNER & Hindi & hin & 11 & 75,827 & 10,851 & 21,657 & 108,335 \\
hr500k & Croatian & hrv & 5 & 17,869 & 2,499 & 4,341 & 24,709 \\
Japanese-GSD & Japanese & jpn & 22 & 7,050 & 507 & 543 & 8,100 \\
KazNERD & Kazakh & kaz & 25 & 90,228 & 11,167 & 11,307 & 112,702 \\
KIND & Italian & ita & 3 & 37,765 & 0 & 7,385 & 45,150 \\
L3Cube MahaNER & Marathi & mar & 7 & 21,493 & 1,499 & 1,998 & 24,990 \\
MasakhaNER & Amharic & amh & 4 & 1,750 & 250 & 500 & 2,500 \\
MasakhaNER & Hausa & hau & 4 & 1,903 & 272 & 545 & 2,720 \\
MasakhaNER & Igbo & ibo & 4 & 2,233 & 319 & 638 & 3,190 \\
MasakhaNER & Kinyarwanda & kin & 4 & 2,110 & 301 & 604 & 3,015 \\
MasakhaNER & Luganda & lug & 4 & 1,402 & 200 & 401 & 2,003 \\
MasakhaNER & Luo & luo & 4 & 644 & 92 & 185 & 921 \\
MasakhaNER & Naija & pcm & 4 & 2,100 & 300 & 600 & 3,000 \\
MasakhaNER & Kiswahili & swa & 4 & 2,104 & 300 & 602 & 3,006 \\
MasakhaNER & Wolof & wol & 4 & 1,871 & 267 & 536 & 2,674 \\
MasakhaNER & Yoruba & yor & 4 & 2,124 & 303 & 608 & 3,035 \\
MasakhaNER 2.0 & Bambara & bam & 4 & 4,462 & 638 & 1,274 & 6,374 \\
MasakhaNER 2.0 & Ghomálá’ & bbj & 4 & 3,384 & 483 & 966 & 4,833 \\
MasakhaNER 2.0 & Éwé & ewe & 4 & 3,505 & 501 & 1,001 & 5,007 \\
MasakhaNER 2.0 & Fon & fon & 4 & 4,343 & 623 & 1,228 & 6,194 \\
MasakhaNER 2.0 & Hausa & hau & 4 & 5,716 & 816 & 1,633 & 8,165 \\
MasakhaNER 2.0 & Igbo & ibo & 4 & 7,634 & 1,090 & 2,181 & 10,905 \\
MasakhaNER 2.0 & Kinyarwanda & kin & 4 & 7,825 & 1,118 & 2,235 & 11,178 \\
MasakhaNER 2.0 & Luganda & lug & 4 & 4,942 & 706 & 1,412 & 7,060 \\
MasakhaNER 2.0 & Luo & luo & 4 & 5,161 & 737 & 1,474 & 7,372 \\
MasakhaNER 2.0 & Mossi & mos & 4 & 4,532 & 648 & 1,294 & 6,474 \\
MasakhaNER 2.0 & Chichewa & nya & 4 & 6,250 & 893 & 1,785 & 8,928 \\
MasakhaNER 2.0 & Naija & pcm & 4 & 5,646 & 806 & 1,613 & 8,065 \\
MasakhaNER 2.0 & chiShona & sna & 4 & 6,207 & 887 & 1,773 & 8,867 \\
MasakhaNER 2.0 & Kiswahili & swa & 4 & 6,593 & 942 & 1,883 & 9,418 \\
MasakhaNER 2.0 & Setswana & tsn & 4 & 3,489 & 499 & 996 & 4,984 \\
MasakhaNER 2.0 & Akan/Twi & twi & 4 & 4,240 & 605 & 1,211 & 6,056 \\
MasakhaNER 2.0 & Wolof & wol & 4 & 4,593 & 656 & 1,312 & 6,561 \\
MasakhaNER 2.0 & isiXhosa & xho & 4 & 5,718 & 817 & 1,633 & 8,168 \\
MasakhaNER 2.0 & Yorùbá & yor & 4 & 6,876 & 983 & 1,964 & 9,823 \\
MasakhaNER 2.0 & Zulu & zul & 4 & 5,848 & 836 & 1,670 & 8,354 \\
NEMO SPMRL & Hebrew & heb & 9 & 4,937 & 500 & 706 & 6,143 \\
NEMO UD & Hebrew & heb & 9 & 5,168 & 484 & 491 & 6,143 \\
NorNE & Norwegian (Nynorsk) & nno & 9 & 14,174 & 1,890 & 1,511 & 17,575 \\
NorNE & Norwegian (Bokmål) & nob & 9 & 15,696 & 2,410 & 1,939 & 20,045 \\
RONEC & Romanian & ron & 15 & 9,000 & 1,330 & 2,000 & 12,330 \\
SLI Galician Corpora & Galician & glg & 4 & 6,483 & 0 & 1,655 & 8,138 \\
ssj500k & Slovenian & slv & 5 & 9,077 & 1,147 & 1,134 & 11,358 \\
ThaiNNER & Thai & tha & 10 & 3,914 & 0 & 980 & 4,894 \\
TurkuNLP & Finnish & fin & 6 & 12,217 & 1,364 & 1,555 & 15,136 \\
Tweebank & English & eng & 4 & 1,639 & 710 & 1,201 & 3,550 \\
UNER Chinese GSD & Mandarin Chinese & cmn & 3 & 3,997 & 500 & 500 & 4,997 \\
UNER Chinese GSDSIMP & Mandarin Chinese & cmn & 3 & 3,997 & 500 & 500 & 4,997 \\
UNER English EWT & English & eng & 3 & 12,543 & 2,001 & 2,077 & 16,621 \\
UNER Maghrebi French-Arabic & Maghrebi Arabic & arq & 3 & 1,003 & 139 & 145 & 1,287 \\
UNER Portuguese-Bosque & Portuguese & por & 3 & 7,018 & 1,172 & 1,167 & 9,357 \\
UNER Slovak SNK & Slovak & slk & 3 & 8,483 & 1,060 & 1,061 & 10,604 \\
UNER Swedish Talkbanken & Swedish & swe & 3 & 4,303 & 504 & 1,219 & 6,026 \\
WikiGoldSK & Slovak & slk & 4 & 4,687 & 669 & 1,340 & 6,696 \\
WNUT17 & English & eng & 6 & 3,394 & 1,009 & 1,287 & 5,690 \\
\midrule
Total &  &  & \multicolumn{1}{l}{} & 624,532 & 76,746 & 130,786 & 832,064 \\
\bottomrule
\end{tabular}
}
\caption{Statistics for corpora included in OpenNER. Language codes are given using the ISO 639-3 standard.}
\label{tab:stats}
\end{table*}

\FloatBarrier

\section*{Limitations}

Despite tremendous efforts to include every eligible dataset, there may be datasets that met our criteria that we missed due to the difficulty of trying to find every single hand-annotated NER dataset in existence.
We hope that our commitment to regular releases will mitigate this limitation.

OpenNER is a collection of existing corpora, and thus it faithfully represents the biases in both the construction of such corpora (i.e. which languages they are created in) and their contents.
Due to the recent release of the MasakhaNER datasets, African languages are overrepresented in OpenNER compared to those of the rest of the world.
OpenNER has substantial coverage of European and African languages but has little coverage of languages spoken in Asia and South America beyond the majority languages spoken in the larger countries.
Unfortunately, part of the cause of underrepresentation of Asian languages is due to many of the NER datasets existing in those languages not being in usable condition (see Appendix~\ref{app:not-included-data}).
Outside of Africa, there is little coverage of indigenous or minority languages due to the limited corpora in existence. 

The majority of the languages included in OpenNER are written in Latin script, but broader script coverage is likely to be a key element of building multilingual NER models.

When mapping to core types, users should be aware that although datasets might be using the same entity type names, these were not annotated using common guidelines and are hence expected to be noisy. 
While this is a limitation, it also presents an opportunity for future work in exploring better mappings from the standardized version of OpenNER to alternative unified types or other approaches to train models to learn from datasets in disparate ontologies.

\section*{Ethical Considerations}

We believe OpenNER will have a positive impact on future multilingual NER research. 
Compared with ``silver-standard'' datasets, OpenNER is high-quality human-annotated data and will bring attention to multilingual gold standard NER datasets that had previously received less attention than some of the larger ``silver-standard'' datasets.

We have undertaken significant efforts to confirm that all datasets we include allow redistribution and were not derived from more restrictive sources that would not allow it.
However, there is always risk that the authors of datasets have misrepresented the restrictions on the data that was annotated, causing us to accidentally redistribute data against the original data owner's wishes.

\section*{Acknowledgments}
This work was primarily supported by the grant \emph{Improving Relevance and Recovery by Extracting Latent Query Structure} by eBay to Brandeis University.
This work was also supported by Brandeis University through internal research funds.

\bibliography{anthology,custom}

\appendix

\section{Appendix}
\label{sec:appendix}

\subsection{Data Formatting Corrections}
\label{sec:formatting-corrections}

hr500k, ssj500k, and NorNE are represented in CoNLL-U Plus format, which does not explicitly include \texttt{O} tags.
We converted these datasets to CoNLL format using SeqScore. 

In the KIND corpus, each token is annotated with just the type name (e.g. \texttt{LOC}). We converted to BIO encoding by prepending all type labels with \text{I-}, and then using SeqScore to convert from IO to BIO encoding.

The L3Cube MahaNER dataset delineates sentence breaks with sentence IDs. We added appropriate newlines. We also standardized the encoding prefixes to be separate from the type name with a dash (e.g. \texttt{BNEO} to \texttt{B-NEO}, \texttt{BLOC} to \texttt{B-LOC}). 

The data for MasakhaNER is taken from commit \texttt{9745180390b3507858ea57f7b1e4f8a944d280fc} due to later commits splitting sentences at an arbitrary maximum length, causing some entity mentions to be split across two sentences.
Two lines in MasakhaNER 2.0 contain only O labels, with no corresponding tokens. We removed these two lines.

RONEC is distributed in JSON format with BIO-encoded labels and tokens as fields. We converted it to CoNLL format.

The ThaiNNER dataset uses BIOES encoding, and uses a nested ontology. We used the top layer of the nested annotation and converted the encoding to BIO. 
ThaiNNER comes with two levels of entity types, coarse-grained and fine-grained. 
We use the coarse-grained types and the top layer of annotation here since the coarse-grained types directly align with most OntoNotes entity types which aligns best with the majority of other corpora we collected. 

CoNLL-02 is distributed using ISO-8859-1 text encoding.
We converted the files to UTF-8.

Unfortunately, the majority of the NER corpora included in OpenNER do not contain document boundary information. 
Sometimes this is due to copyright limitations in the original dataset, but often it was due to poor data preparation practices and many old NER data preparation scripts removing document boundaries.
OpenNER retains any document boundaries if it was included in the source data. 
CoNLL-02 Dutch, the UNER corpora (except Maghrebi Arabic), TurkuNLP, hr500k, and ssj500k retain document boundaries. 
OpenNER uses the convention introduced by CoNLL-02 of marking document boundaries with the sentence \texttt{-DOCSTART- O}.

\subsection{Label Repair}
\label{sec:repair}
Invalid label sequences in datasets need to be reviewed manually to ensure the problem is just an issue with invalid transitions and not an annotation error. 
Once reviewed manually, many errors can be repaired automatically.
For example, when SeqScore encounters the label sequence \texttt{O I-PER I-PER O} in a dataset that is supposed to be BIO encoded, it is repaired to \texttt{O B-PER I-PER O} using the same approach taken in the original \texttt{conlleval} script.

In most cases, automatic repair using SeqScore is possible after brief manual review. 
This approach corrected 108 errors across the included datasets.
In 32 cases for SLI Galician, manual repairs were performed.
While most of these could be repaired using the \texttt{conlleval} approach, there were 14 which would have been incorrectly labeled using an automatic repair.

\begin{table*}[tbh]
\centering
\footnotesize
\vspace{-0.2in}
\adjustbox{max height=0.50\textheight}{%
\begin{tabular}{lllll}
\toprule
Dataset & Lang. & mBERT & XLM-R & Glot500 \\
\midrule
AnCora & cat & 76.93 $^{\pm0.20}$ & \textbf{86.71} $^{\pm0.12}$ & 85.82 $^{\pm0.13}$ \\
AnCora & spa & 86.58 $^{\pm0.07}$ & \textbf{91.89} $^{\pm0.05}$ & 91.78 $^{\pm0.08}$ \\
AQMAR & ara & 73.00 $^{\pm0.15}$ & \textbf{75.61} $^{\pm0.27}$ & 73.53 $^{\pm0.51}$ \\
ArmanPersoNER & fas & 80.44 $^{\pm0.09}$ & \textbf{82.25} $^{\pm0.12}$ & 81.32 $^{\pm0.12}$ \\
BarNER & bar & 47.35 $^{\pm0.42}$ & 60.53 $^{\pm0.32}$ & \textbf{61.97} $^{\pm0.62}$ \\
CoNLL-02 & nld & 84.54 $^{\pm0.13}$ & \textbf{89.83} $^{\pm0.09}$ & 89.29 $^{\pm0.13}$ \\
CoNLL-02 & spa & 81.85 $^{\pm0.13}$ & \textbf{87.36} $^{\pm0.13}$ & 86.72 $^{\pm0.12}$ \\
DaNE & dan & 77.01 $^{\pm0.31}$ & \textbf{84.09} $^{\pm0.26}$ & 81.84 $^{\pm0.28}$ \\
EIEC & eus & 69.21 $^{\pm0.24}$ & \textbf{82.20} $^{\pm0.27}$ & 78.91 $^{\pm0.33}$ \\
elNER & ell & 88.35 $^{\pm0.09}$ & \textbf{91.99} $^{\pm0.08}$ & 91.45 $^{\pm0.08}$ \\
EverestNER & nep & 85.09 $^{\pm0.12}$ & 90.18 $^{\pm0.11}$ & \textbf{90.30} $^{\pm0.08}$ \\
GermEval & deu & 81.77 $^{\pm0.12}$ & \textbf{84.98} $^{\pm0.11}$ & 84.65 $^{\pm0.1}$ \\
HiNER & hin & 88.99 $^{\pm0.01}$ & \textbf{90.00} $^{\pm0.02}$ & 88.69 $^{\pm1.11}$ \\
hr500k & hrv & 75.54 $^{\pm0.16}$ & \textbf{87.03} $^{\pm0.13}$ & 86.78 $^{\pm0.09}$ \\
Japanese GSD & jpn & 81.08 $^{\pm0.21}$ & \textbf{83.46} $^{\pm0.24}$ & 80.87 $^{\pm0.23}$ \\
KazNERD & kaz & 95.48 $^{\pm0.04}$ & \textbf{96.38} $^{\pm0.04}$ & 86.37 $^{\pm9.6}$ \\
KIND & ita & 83.13 $^{\pm0.06}$ & \textbf{86.62} $^{\pm0.12}$ & \textbf{86.62} $^{\pm0.07}$ \\
L3Cube MahaNER & mar & 81.10 $^{\pm0.14}$ & 83.05 $^{\pm0.15}$ & \textbf{83.31} $^{\pm0.16}$ \\
MasakhaNER & amh & 00.00 $^{\pm0.00}$ & \textbf{70.33} $^{\pm0.45}$ & 57.05 $^{\pm9.52}$ \\
MasakhaNER & hau & 81.60 $^{\pm0.37}$ & \textbf{89.03} $^{\pm0.21}$ & 88.66 $^{\pm0.23}$ \\
MasakhaNER & ibo & 76.20 $^{\pm0.31}$ & 83.66 $^{\pm0.33}$ & \textbf{86.33} $^{\pm0.28}$ \\
MasakhaNER & kin & 61.40 $^{\pm0.64}$ & 71.65 $^{\pm0.47}$ & \textbf{76.07} $^{\pm0.18}$ \\
MasakhaNER & lug & 72.38 $^{\pm0.24}$ & 77.74 $^{\pm0.41}$ & \textbf{83.00} $^{\pm0.27}$ \\
MasakhaNER & luo & 61.03 $^{\pm1.10}$ & \textbf{71.01} $^{\pm0.65}$ & 54.42 $^{\pm2.77}$ \\
MasakhaNER & pcm & 80.02 $^{\pm0.26}$ & 86.87 $^{\pm0.25}$ & \textbf{88.90} $^{\pm0.19}$ \\
MasakhaNER & swa & 82.24 $^{\pm0.25}$ & \textbf{86.80} $^{\pm0.22}$ & 85.67 $^{\pm0.26}$ \\
MasakhaNER & wol & 44.36 $^{\pm4.94}$ & 62.03 $^{\pm0.49}$ & \textbf{65.24} $^{\pm0.83}$ \\
MasakhaNER & yor & 72.88 $^{\pm0.25}$ & 75.10 $^{\pm0.44}$ & \textbf{81.17} $^{\pm0.37}$ \\
MasakhaNER 2.0 & bam & 77.65 $^{\pm0.24}$ & 78.86 $^{\pm0.35}$ & \textbf{79.70} $^{\pm0.22}$ \\
MasakhaNER 2.0 & bbj & 69.57 $^{\pm0.36}$ & \textbf{71.66} $^{\pm0.51}$ & 68.52 $^{\pm0.44}$ \\
MasakhaNER 2.0 & ewe & 81.12 $^{\pm0.14}$ & 87.83 $^{\pm0.19}$ & \textbf{89.17} $^{\pm0.18}$ \\
MasakhaNER 2.0 & fon & 77.81 $^{\pm0.27}$ & 80.85 $^{\pm0.26}$ & \textbf{82.10} $^{\pm0.18}$ \\
MasakhaNER 2.0 & hau & 71.44 $^{\pm0.28}$ & \textbf{83.77} $^{\pm0.22}$ & 83.62 $^{\pm0.10}$ \\
MasakhaNER 2.0 & ibo & 81.60 $^{\pm0.20}$ & 86.49 $^{\pm0.33}$ & \textbf{89.39} $^{\pm0.36}$ \\
MasakhaNER 2.0 & kin & 77.86 $^{\pm0.22}$ & 81.85 $^{\pm0.26}$ & \textbf{86.08} $^{\pm0.12}$ \\
MasakhaNER 2.0 & lug & 84.06 $^{\pm0.17}$ & 86.23 $^{\pm0.19}$ & \textbf{88.51} $^{\pm0.13}$ \\
MasakhaNER 2.0 & luo & 74.19 $^{\pm0.21}$ & 79.31 $^{\pm0.19}$ & \textbf{81.87} $^{\pm0.2}$ \\
MasakhaNER 2.0 & mos & 60.30 $^{\pm0.28}$ & 73.29 $^{\pm0.35}$ & \textbf{76.03} $^{\pm0.28}$ \\
MasakhaNER 2.0 & nya & 85.66 $^{\pm0.22}$ & 89.42 $^{\pm0.09}$ & \textbf{91.64} $^{\pm0.09}$ \\
MasakhaNER 2.0 & pcm & 84.00 $^{\pm0.14}$ & 88.34 $^{\pm0.15}$ & \textbf{89.24} $^{\pm0.10}$ \\
MasakhaNER 2.0 & sna & 89.49 $^{\pm0.11}$ & 92.93 $^{\pm0.20}$ & \textbf{95.05} $^{\pm0.09}$ \\
MasakhaNER 2.0 & swa & 89.75 $^{\pm0.07}$ & 91.86 $^{\pm0.07}$ & \textbf{92.02} $^{\pm0.06}$ \\
MasakhaNER 2.0 & tsn & 82.08 $^{\pm0.21}$ & 85.15 $^{\pm0.29}$ & \textbf{87.79} $^{\pm0.17}$ \\
MasakhaNER 2.0 & twi & 72.08 $^{\pm0.23}$ & 77.49 $^{\pm0.39}$ & \textbf{80.16} $^{\pm0.36}$ \\
MasakhaNER 2.0 & wol & 77.43 $^{\pm0.16}$ & 81.00 $^{\pm0.55}$ & \textbf{85.48} $^{\pm0.22}$ \\
MasakhaNER 2.0 & xho & 78.31 $^{\pm0.18}$ & 86.33 $^{\pm0.07}$ & \textbf{87.91} $^{\pm0.17}$ \\
MasakhaNER 2.0 & yor & 82.00 $^{\pm0.18}$ & 85.30 $^{\pm0.22}$ & \textbf{86.76} $^{\pm0.36}$ \\
MasakhaNER 2.0 & zul & 72.80 $^{\pm0.19}$ & 83.65 $^{\pm0.33}$ & \textbf{86.53} $^{\pm0.23}$ \\
NEMO SPMRL & heb & 76.68 $^{\pm0.35}$ & \textbf{80.02} $^{\pm0.43}$ & 76.60 $^{\pm0.56}$ \\
NEMO UD & heb & 73.80 $^{\pm0.28}$ & \textbf{76.44} $^{\pm0.49}$ & 74.40 $^{\pm0.38}$ \\
NorNE & nno & 78.53 $^{\pm0.38}$ & 85.30 $^{\pm0.32}$ & \textbf{85.48} $^{\pm0.25}$ \\
NorNE & nob & 74.26 $^{\pm0.27}$ & \textbf{87.14} $^{\pm0.21}$ & 85.40 $^{\pm0.23}$ \\
RONEC & ron & 86.14 $^{\pm0.04}$ & \textbf{88.65} $^{\pm0.05}$ & 87.82 $^{\pm0.07}$ \\
SLI Galician Corpora & glg & 76.16 $^{\pm0.16}$ & \textbf{87.08} $^{\pm0.24}$ & 85.94 $^{\pm0.25}$ \\
ssj500k & slv & 51.73 $^{\pm0.65}$ & \textbf{60.79} $^{\pm0.42}$ & 55.51 $^{\pm6.17}$ \\
ThaiNNER & tha & 64.34 $^{\pm0.03}$ & 71.94 $^{\pm0.18}$ & \textbf{72.19} $^{\pm0.05}$ \\
TurkuNLP & fin & 78.04 $^{\pm0.22}$ & \textbf{87.04} $^{\pm0.18}$ & 86.08 $^{\pm0.25}$ \\
Tweebank & eng & 52.59 $^{\pm0.32}$ & \textbf{60.93} $^{\pm2.06}$ & 50.45 $^{\pm2.79}$ \\
UNER Chinese GSD & cmn & \textbf{87.15} $^{\pm0.21}$ & 85.10 $^{\pm0.29}$ & 86.37 $^{\pm0.21}$ \\
UNER Chinese GSDSIMP & cmn & \textbf{87.52} $^{\pm0.21}$ & 84.75 $^{\pm0.39}$ & 77.29 $^{\pm8.59}$ \\
UNER English EWT & eng & 75.46 $^{\pm0.22}$ & 80.61 $^{\pm0.30}$ & \textbf{81.53} $^{\pm0.24}$ \\
UNER Maghrebi Arabic & arq & \textbf{81.30} $^{\pm0.35}$ & 73.30 $^{\pm1.66}$ & 65.28 $^{\pm1.30}$ \\
UNER Portuguese-Bosque & por & 80.73 $^{\pm0.23}$ & \textbf{88.48} $^{\pm0.22}$ & 87.74 $^{\pm0.20}$ \\
UNER Slovak SNK & slk & 55.83 $^{\pm0.94}$ & \textbf{77.44} $^{\pm0.71}$ & 73.82 $^{\pm0.47}$ \\
UNER Swedish Talkbanken & swe & 60.33 $^{\pm10.08}$ & \textbf{84.42} $^{\pm0.76}$ & 72.01 $^{\pm8.28}$ \\
WikiGoldSK & slk & 84.98 $^{\pm0.18}$ & \textbf{90.45} $^{\pm0.31}$ & 89.70 $^{\pm0.18}$\\
WNUT17 & eng & 33.07 $^{\pm0.50}$ & \textbf{50.15} $^{\pm0.39}$ & 46.45 $^{\pm0.45}$\\
\midrule
Mean & & 74.42 $^{\pm{1.78}}$ & 81.79 $^{\pm{1.08}}$ & 80.96 $^{\pm{1.30}}$ \\
\bottomrule
\end{tabular}
}
\caption{Mean F1 $\pm$ standard error for individual (per language-dataset) models.}
\label{tab:indiv-results}
\end{table*}

\begin{table*}[tbh]
\centering
\footnotesize
\vspace{-0.2in}
\adjustbox{max height=0.50\textheight}{
\begin{tabular}{llllllll}
\toprule
 &  & \multicolumn{3}{c}{Individual} & \multicolumn{3}{c}{Multilingual} \\
\cmidrule(lr){3-5} \cmidrule(lr){6-8}
Dataset & Lang. & mBERT & XLM-R & Glot500 & mBERT & XLM-R & Glot500 \\
\midrule

AnCora                                 & cat   & 80.59 $^{\pm0.13}$ & 88.71 $^{\pm0.12}$ & 87.74 $^{\pm0.22}$  & 80.87 $^{\pm0.15}$ & \textbf{88.31} $^{\pm0.15}$ & 88.27 $^{\pm0.11}$ \\
AnCora                                 & spa   & 86.13 $^{\pm0.08}$ & \textbf{91.92} $^{\pm0.07}$ & 91.74 $^{\pm0.05}$  & 85.90 $^{\pm0.16}$  & 91.46 $^{\pm0.11}$ & 91.37 $^{\pm0.08}$ \\
AQMAR                                  & ara   & 78.81 $^{\pm0.25}$ & \textbf{80.89} $^{\pm0.39}$ & 63.93 $^{\pm10.66}$ & 71.02 $^{\pm0.17}$ & 76.33 $^{\pm0.24}$ & 77.37 $^{\pm0.25}$ \\
ArmanPersoNER                          & fas   & 82.68 $^{\pm0.11}$ & \textbf{84.66} $^{\pm0.06}$ & 83.62 $^{\pm0.06}$  & 81.08 $^{\pm0.07}$ & 83.06 $^{\pm0.07}$ & 82.81 $^{\pm0.12}$ \\
BarNER                                 & bar   & 51.54 $^{\pm0.48}$ & 68.47 $^{\pm0.65}$ & 71.26 $^{\pm0.86}$  & 69.94 $^{\pm0.65}$ & 74.35 $^{\pm0.53}$ & \textbf{77.70} $^{\pm0.54}$  \\
CONLL02                                & nld   & 85.96 $^{\pm0.13}$ & 90.61 $^{\pm0.12}$ & 90.11 $^{\pm0.14}$  & 85.60 $^{\pm0.08}$  & 90.56 $^{\pm0.12}$ & \textbf{90.79} $^{\pm0.10}$  \\
CONLL02                                & spa   & 84.42 $^{\pm0.07}$ & \textbf{89.05} $^{\pm0.13}$ & 88.31 $^{\pm0.14}$  & 84.68 $^{\pm0.18}$ & 88.40 $^{\pm0.10}$   & 88.50 $^{\pm0.11}$  \\
DaNE                                   & dan   & 78.41 $^{\pm0.33}$ & \textbf{85.32} $^{\pm0.22}$ & 83.09 $^{\pm0.52}$  & 77.79 $^{\pm0.40}$  & 84.02 $^{\pm0.21}$ & 82.82 $^{\pm0.29}$ \\
EIEC                                   & eus   & 70.96 $^{\pm0.27}$ & \textbf{82.66} $^{\pm0.29}$ & 80.60 $^{\pm0.32}$   & 68.96 $^{\pm0.24}$ & 80.96 $^{\pm0.21}$ & 79.51 $^{\pm0.29}$ \\
elNER                                  & ell   & 86.40 $^{\pm0.12}$  & 91.48 $^{\pm0.04}$ & 91.32 $^{\pm0.08}$  & 86.39 $^{\pm0.07}$ & \textbf{91.57} $^{\pm0.06}$ & \textbf{91.57} $^{\pm0.11}$ \\
EverestNER                             & nep   & 84.57 $^{\pm0.12}$ & 90.41 $^{\pm0.12}$ & \textbf{90.55} $^{\pm0.10}$   & 85.02 $^{\pm0.14}$ & 89.91 $^{\pm0.11}$ & 89.90 $^{\pm0.16}$  \\
GermEval                               & deu   & 84.61 $^{\pm0.10}$  & 87.57 $^{\pm0.09}$ & \textbf{87.61} $^{\pm0.06}$  & 83.24 $^{\pm0.11}$ & 86.11 $^{\pm0.09}$ & 86.57 $^{\pm0.09}$ \\
HiNER                                  & hin   & 91.87 $^{\pm0.02}$ & \textbf{92.73} $^{\pm0.01}$ & 92.06 $^{\pm0.66}$  & 91.21 $^{\pm0.02}$ & 92.35 $^{\pm0.01}$ & 92.34 $^{\pm0.02}$ \\
hr500k                                 & hrv   & 78.32 $^{\pm0.12}$ & 88.47 $^{\pm0.13}$ & \textbf{88.49} $^{\pm0.09}$  & 78.62 $^{\pm0.10}$  & 87.97 $^{\pm0.13}$ & 88.18 $^{\pm0.13}$ \\
Japanese GSD                          & jpn   & 79.04 $^{\pm0.52}$ & \textbf{80.75} $^{\pm0.62}$ & 63.36 $^{\pm10.57}$ & 74.63 $^{\pm0.50}$  & 78.01 $^{\pm0.28}$ & 78.75 $^{\pm0.34}$ \\
KazNERD                                & kaz   & \textbf{91.80} $^{\pm0.13}$  & 85.20 $^{\pm8.89}$  & 83.65 $^{\pm9.32}$  & 91.40 $^{\pm0.10}$   & 94.11 $^{\pm0.09}$ & 94.29 $^{\pm0.07}$ \\
KIND                                   & ita   & 83.16 $^{\pm0.09}$ & 86.56 $^{\pm0.07}$ & \textbf{86.65} $^{\pm0.09}$  & 81.57 $^{\pm0.12}$ & 85.98 $^{\pm0.08}$ & 86.30 $^{\pm0.13}$  \\
L3Cube MahaNER                        & mar   & 79.14 $^{\pm0.18}$ & \textbf{82.75} $^{\pm0.16}$ & 66.07 $^{\pm11.01}$ & 78.34 $^{\pm0.16}$ & 81.70 $^{\pm0.19}$  & 82.23 $^{\pm0.21}$ \\
MasakhaNER                             & amh   & 00.00 $^{\pm0.00}$        & 69.40 $^{\pm0.78}$  & \textbf{71.25} $^{\pm0.61}$  & 00.00 $^{\pm0.00}$        & 70.52 $^{\pm0.51}$ & 70.95 $^{\pm0.31}$ \\
MasakhaNER                             & hau   & 82.45 $^{\pm0.18}$ & \textbf{90.62} $^{\pm0.20}$  & 89.95 $^{\pm0.14}$  & 85.13 $^{\pm0.24}$ & 89.13 $^{\pm0.18}$ & 90.08 $^{\pm0.17}$ \\
MasakhaNER                             & ibo   & 76.80 $^{\pm0.26}$  & 85.85 $^{\pm0.31}$ & 88.06 $^{\pm0.46}$  & 84.66 $^{\pm0.23}$ & 88.19 $^{\pm0.20}$  & \textbf{90.55} $^{\pm0.21}$ \\
MasakhaNER                             & kin   & 60.00 $^{\pm0.67}$    & 72.87 $^{\pm0.41}$ & 76.73 $^{\pm0.30}$   & 76.29 $^{\pm0.22}$ & 79.83 $^{\pm0.20}$  & \textbf{83.07} $^{\pm0.11}$ \\
MasakhaNER                             & lug   & 72.20 $^{\pm0.40}$   & 79.33 $^{\pm0.43}$ & 83.79 $^{\pm0.57}$  & 76.45 $^{\pm0.24}$ & 81.51 $^{\pm0.24}$ & \textbf{85.56} $^{\pm0.18}$ \\
MasakhaNER                             & luo   & 61.29 $^{\pm0.70}$  & 72.92 $^{\pm0.70}$  & 60.58 $^{\pm4.33}$  & 82.20 $^{\pm0.54}$  & 82.35 $^{\pm0.35}$ & \textbf{82.82} $^{\pm0.26}$ \\
MasakhaNER                             & pcm   & 78.13 $^{\pm0.32}$ & 85.82 $^{\pm0.57}$ & 88.62 $^{\pm0.27}$  & 86.51 $^{\pm0.24}$ & 90.21 $^{\pm0.27}$ & \textbf{91.38} $^{\pm0.26}$ \\
MasakhaNER                             & swa   & 81.88 $^{\pm0.23}$ & 87.37 $^{\pm0.16}$ & 86.74 $^{\pm0.16}$  & 87.32 $^{\pm0.17}$ & 89.29 $^{\pm0.22}$ & \textbf{89.70} $^{\pm0.12}$  \\
MasakhaNER                             & wol   & 53.68 $^{\pm0.41}$ & 67.84 $^{\pm0.48}$ & 69.29 $^{\pm1.31}$  & 69.09 $^{\pm0.61}$ & 72.40 $^{\pm0.42}$  & \textbf{75.55} $^{\pm0.32}$ \\
MasakhaNER                             & yor   & 72.69 $^{\pm0.20}$  & 76.39 $^{\pm0.36}$ & 80.36 $^{\pm0.94}$  & 82.97 $^{\pm0.16}$ & 85.10 $^{\pm0.15}$  & \textbf{88.43} $^{\pm0.18}$ \\
MasakhaNER 2.0                        & bam   & 76.21 $^{\pm0.27}$ & 77.82 $^{\pm0.26}$ & 78.10 $^{\pm0.47}$   & 75.24 $^{\pm0.27}$ & 80.14 $^{\pm0.18}$ & \textbf{81.07} $^{\pm0.25}$ \\
MasakhaNER 2.0                        & bbj   & 68.52 $^{\pm0.38}$ & 70.69 $^{\pm0.29}$ & 69.26 $^{\pm0.34}$  & 72.42 $^{\pm0.33}$ & 72.93 $^{\pm0.31}$ & \textbf{73.32} $^{\pm0.22}$ \\
MasakhaNER 2.0                        & ewe   & 84.41 $^{\pm0.14}$ & 88.96 $^{\pm0.15}$ & 90.69 $^{\pm0.15}$  & 88.60 $^{\pm0.10}$   & 90.78 $^{\pm0.12}$ & \textbf{91.68} $^{\pm0.12}$ \\
MasakhaNER 2.0                        & fon   & 80.00 $^{\pm0.35}$    & 83.60 $^{\pm0.16}$  & 84.68 $^{\pm0.23}$  & 82.02 $^{\pm0.25}$ & 84.74 $^{\pm0.30}$  & \textbf{86.70} $^{\pm0.37}$  \\
MasakhaNER 2.0                        & hau   & 71.78 $^{\pm0.37}$ & 84.99 $^{\pm0.20}$  & 85.22 $^{\pm0.23}$  & 80.64 $^{\pm0.13}$ & 84.20 $^{\pm0.23}$  & \textbf{85.47} $^{\pm0.21}$ \\
MasakhaNER 2.0                        & ibo   & 83.30 $^{\pm0.19}$  & 89.70 $^{\pm0.28}$  & 90.31 $^{\pm0.30}$   & 88.71 $^{\pm0.24}$ & 93.78 $^{\pm0.06}$ & \textbf{94.09} $^{\pm0.54}$ \\
MasakhaNER 2.0                        & kin   & 78.87 $^{\pm0.17}$ & 84.72 $^{\pm0.24}$ & \textbf{88.38} $^{\pm0.16}$  & 84.65 $^{\pm0.16}$ & 86.08 $^{\pm0.12}$ & 88.16 $^{\pm0.10}$  \\
MasakhaNER 2.0                        & lug   & 86.94 $^{\pm0.21}$ & 89.92 $^{\pm0.10}$  & 91.87 $^{\pm0.08}$  & 88.10 $^{\pm0.08}$  & 90.79 $^{\pm0.11}$ & \textbf{92.12} $^{\pm0.08}$ \\
MasakhaNER 2.0                        & luo   & 74.38 $^{\pm0.18}$ & 80.00 $^{\pm0.20}$     & \textbf{82.71} $^{\pm0.20}$   & 77.01 $^{\pm0.21}$ & 80.99 $^{\pm0.24}$ & 82.40 $^{\pm0.15}$  \\
MasakhaNER 2.0                        & mos   & 61.07 $^{\pm0.42}$ & 75.94 $^{\pm0.42}$ & 76.62 $^{\pm0.44}$  & 65.29 $^{\pm0.33}$ & \textbf{76.72} $^{\pm0.21}$ & 75.86 $^{\pm0.38}$ \\
MasakhaNER 2.0                        & nya   & 86.75 $^{\pm0.22}$ & 91.07 $^{\pm0.11}$ & \textbf{93.22} $^{\pm0.08}$  & 87.67 $^{\pm0.09}$ & 90.72 $^{\pm0.05}$ & 92.80 $^{\pm0.09}$  \\
MasakhaNER 2.0                        & pcm   & 82.79 $^{\pm0.26}$ & 87.92 $^{\pm0.10}$  & 88.82 $^{\pm0.17}$  & 84.37 $^{\pm0.18}$ & 87.78 $^{\pm0.12}$ & \textbf{88.96} $^{\pm0.09}$ \\
MasakhaNER 2.0                        & sna   & 89.61 $^{\pm0.15}$ & 93.60 $^{\pm0.15}$  & 95.51 $^{\pm0.08}$  & 90.23 $^{\pm0.12}$ & 94.58 $^{\pm0.07}$ & \textbf{95.54} $^{\pm0.07}$ \\
MasakhaNER 2.0                        & swa   & 91.93 $^{\pm0.12}$ & 94.27 $^{\pm0.07}$ & 94.17 $^{\pm0.06}$  & 92.49 $^{\pm0.08}$ & 94.35 $^{\pm0.08}$ & \textbf{94.48} $^{\pm0.07}$ \\
MasakhaNER 2.0                        & tsn   & 83.47 $^{\pm0.42}$ & 87.83 $^{\pm0.28}$ & 89.41 $^{\pm0.25}$  & 85.28 $^{\pm0.17}$ & 87.22 $^{\pm0.23}$ & \textbf{89.58} $^{\pm0.15}$ \\
MasakhaNER 2.0                        & twi   & 74.61 $^{\pm0.24}$ & 81.06 $^{\pm0.37}$ & 82.46 $^{\pm0.19}$  & 75.09 $^{\pm0.34}$ & 81.90 $^{\pm0.30}$ & \textbf{83.70} $^{\pm0.33}$  \\
MasakhaNER 2.0                        & wol   & 79.26 $^{\pm0.20}$  & 82.47 $^{\pm0.37}$ & 86.35 $^{\pm0.20}$   & 81.68 $^{\pm0.19}$ & 86.17 $^{\pm0.24}$ & \textbf{87.93} $^{\pm0.15}$ \\
MasakhaNER 2.0                        & xho   & 79.73 $^{\pm0.14}$ & 88.94 $^{\pm0.21}$ & 89.95 $^{\pm0.11}$  & 81.57 $^{\pm0.13}$ & 89.25 $^{\pm0.11}$ & \textbf{90.66} $^{\pm0.08}$ \\
MasakhaNER 2.0                        & yor   & 81.97 $^{\pm0.19}$ & 86.77 $^{\pm0.18}$ & 88.32 $^{\pm0.29}$  & 82.18 $^{\pm0.14}$ & 87.24 $^{\pm0.14}$ & \textbf{88.65} $^{\pm0.15}$ \\
MasakhaNER 2.0                        & zul   & 71.73 $^{\pm0.29}$ & 84.13 $^{\pm0.24}$ & 86.64 $^{\pm0.22}$  & 76.18 $^{\pm0.12}$ & 86.52 $^{\pm0.16}$ & \textbf{89.22} $^{\pm0.19}$ \\
NEMO SPMRL                            & heb   & 79.76 $^{\pm0.55}$ & 81.38 $^{\pm0.23}$ & 78.20 $^{\pm0.42}$   & 86.45 $^{\pm0.15}$ & 88.16 $^{\pm0.14}$ & \textbf{88.26} $^{\pm0.14}$ \\
NEMO UD                               & heb   & 76.36 $^{\pm0.48}$ & 80.28 $^{\pm0.30}$  & \textbf{76.82} $^{\pm0.51}$  & 73.24 $^{\pm0.30}$  & 76.56 $^{\pm0.24}$ & \textbf{76.82} $^{\pm0.26}$ \\
NorNE                                  & nno   & 80.70 $^{\pm0.23}$  & 89.70 $^{\pm0.26}$  & 89.74 $^{\pm0.17}$  & 81.48 $^{\pm0.34}$ & 90.55 $^{\pm0.24}$ & \textbf{91.27} $^{\pm0.23}$ \\
NorNE                                  & nob   & 73.27 $^{\pm0.32}$ & \textbf{88.57} $^{\pm0.23}$ & 77.01 $^{\pm8.56}$  & 76.10 $^{\pm0.45}$  & 88.40 $^{\pm0.17}$  & 87.84 $^{\pm0.18}$ \\
RONEC                                  & ron   & 83.90 $^{\pm0.09}$  & 87.43 $^{\pm0.07}$ & 86.27 $^{\pm0.11}$  & 84.06 $^{\pm0.09}$ & \textbf{87.61} $^{\pm0.06}$ & 87.38 $^{\pm0.06}$ \\
SLI Galician Corpora                 & glg   & 79.43 $^{\pm0.21}$ & 88.42 $^{\pm0.14}$ & 87.70 $^{\pm0.18}$   & 79.94 $^{\pm0.26}$ & \textbf{89.03} $^{\pm0.16}$ & 88.94 $^{\pm0.16}$ \\
ssj500k                                & slv   & 54.26 $^{\pm0.59}$ & 63.45 $^{\pm0.53}$ & 64.14 $^{\pm0.37}$  & 56.80 $^{\pm0.42}$  & \textbf{64.63} $^{\pm0.18}$ & 64.00 $^{\pm0.32}$    \\
ThaiNNER                               & tha   & 63.75 $^{\pm0.08}$ & \textbf{72.79} $^{\pm0.08}$ & 72.75 $^{\pm0.08}$  & 63.23 $^{\pm0.06}$ & 72.29 $^{\pm0.09}$ & 72.00 $^{\pm0.07}$    \\
TurkuNLP                               & fin   & 77.42 $^{\pm0.27}$ & \textbf{88.22} $^{\pm0.28}$ & 86.04 $^{\pm0.37}$  & 77.03 $^{\pm0.33}$ & 86.18 $^{\pm0.20}$  & 85.84 $^{\pm0.23}$ \\
Tweebank                               & eng   & 57.88 $^{\pm0.59}$ & \textbf{70.82} $^{\pm0.27}$ & 62.04 $^{\pm2.47}$  & 63.12 $^{\pm0.26}$ & 69.41 $^{\pm0.36}$ & 70.40 $^{\pm0.59}$  \\
UNER Chinese GSD                     & cmn   & \textbf{87.40} $^{\pm0.15}$  & 85.13 $^{\pm0.17}$ & 83.33 $^{\pm2.69}$  & 83.75 $^{\pm0.40}$  & 85.22 $^{\pm0.31}$ & 85.34 $^{\pm0.24}$ \\
UNER Chinese GSDSIMP                 & cmn   & \textbf{87.31} $^{\pm0.16}$ & 85.52 $^{\pm0.20}$  & 85.52 $^{\pm0.20}$   & 83.59 $^{\pm0.35}$ & 85.28 $^{\pm0.30}$  & 85.56 $^{\pm0.22}$ \\
UNER English EWT                     & eng   & 75.25 $^{\pm0.20}$  & 80.96 $^{\pm0.23}$ & \textbf{81.31} $^{\pm0.10}$   & 75.25 $^{\pm0.28}$ & 78.05 $^{\pm0.30}$  & 79.19 $^{\pm0.26}$ \\
UNER Maghrebi Arabic & arq   & \textbf{81.32} $^{\pm0.39}$ & 75.78 $^{\pm0.32}$ & 65.09 $^{\pm1.67}$  & 79.40 $^{\pm0.48}$  & 77.68 $^{\pm0.56}$ & 78.16 $^{\pm0.45}$ \\
UNER Portuguese-Bosque                & por   & 80.90 $^{\pm0.28}$  & \textbf{88.77} $^{\pm0.20}$  & 88.12 $^{\pm0.22}$  & 81.18 $^{\pm0.23}$ & 88.21 $^{\pm0.15}$ & 87.59 $^{\pm0.15}$ \\
UNER Slovak SNK                      & slk   & 55.54 $^{\pm0.56}$ & 76.70 $^{\pm0.40}$   & 74.46 $^{\pm0.54}$  & 70.23 $^{\pm0.23}$ & \textbf{82.80} $^{\pm0.16}$  & 82.40 $^{\pm0.27}$  \\
UNER Swedish Talkbanken              & swe   & 67.34 $^{\pm6.44}$ & 76.88 $^{\pm8.56}$ & 79.79 $^{\pm3.20}$   & 78.65 $^{\pm0.43}$ & \textbf{86.49} $^{\pm0.54}$ & 84.49 $^{\pm0.67}$ \\
WikiGoldSK                             & slk   & 85.91 $^{\pm0.19}$ & \textbf{92.25} $^{\pm0.15}$ & 90.94 $^{\pm0.13}$  & 83.43 $^{\pm0.13}$ & 90.68 $^{\pm0.18}$ & 90.65 $^{\pm0.10}$  \\
WNUT17                                 & eng   & 38.70 $^{\pm0.50}$   & \textbf{53.85} $^{\pm0.51}$ & 53.02 $^{\pm0.62}$  & 45.23 $^{\pm0.22}$ & 52.61 $^{\pm0.44}$ & 51.86 $^{\pm0.28}$ 
\\ 
\midrule
Mean &  & 75.69$^{\pm1.78}$ & 83.10$^{\pm1.08}$ & 82.10 $^{\pm1.29}$ & 78.33$^{\pm{1.57}}$ & 84.18$^{\pm{0.93}}$ & \textbf{84.89}$^{\pm{0.94}}$\\
\bottomrule

\end{tabular}
}
\centering
\caption{Mean F1 $\pm$ standard error for individual and multilingual models on core types (PER, LOC, ORG).}
\label{tab:multi-results}
\end{table*}

\begin{table*}[tb]
\centering 
\footnotesize
\vspace{-0.2in}
\adjustbox{max height=0.5\textheight}{
\begin{tabular}{llrr}
\toprule
Dataset                             & Language & Aya-Expanse 32b     & QwQ 32b Preview     \\ \midrule
AnCora                              & cat      & \textbf{74.47} $^{\pm 0.61}$ & 74.19 $^{\pm 0.68}$ \\
AnCora                              & spa      & \textbf{71.28} $^{\pm 1.18}$ & 71.23 $^{\pm 1.13}$ \\
AQMAR                               & ara      & 44.80 $^{\pm 1.94}$  & \textbf{45.22} $^{\pm 2.06}$ \\
ArmanPersoNER                       & fas      & 50.94 $^{\pm 0.44}$  & \textbf{53.23} $^{\pm 1.35}$ \\
BarNER                              & bar      & \textbf{57.04} $^{\pm 2.45}$ & \textbf{57.04} $^{\pm 2.45}$ \\
CoNLL-02                             & nld      & \textbf{72.36} $^{\pm 1.18}$ & 71.77 $^{\pm 0.66}$ \\
CoNLL-02                             & spa      & 74.08 $^{\pm 1.29}$ & \textbf{74.45} $^{\pm 1.63}$ \\
DaNE                                & dan      & \textbf{73.02} $^{\pm 1.96}$ & \textbf{73.02} $^{\pm 1.96}$ \\
EIEC                                & eus      & \textbf{65.03} $^{\pm 0.65}$ & \textbf{65.03} $^{\pm 0.65}$ \\
elNER                               & ell      & 58.95 $^{\pm 2.88}$ & \textbf{62.11} $^{\pm 0.31}$ \\
EverestNER                          & nep      & \textbf{57.53} $^{\pm 0.50}$  & \textbf{57.53} $^{\pm 0.50}$  \\
GermEval                            & deu      & \textbf{69.40} $^{\pm 0.84}$  & \textbf{69.40} $^{\pm 0.84}$  \\
HiNER                               & hin      & \textbf{61.27} $^{\pm 5.71}$ & \textbf{61.27} $^{\pm 5.71}$ \\
hr500k                              & hrv      & \textbf{69.32} $^{\pm 0.99}$ & \textbf{69.32} $^{\pm 0.99}$ \\
Japanese GSD                        & jpn      & \textbf{43.45} $^{\pm 1.23}$ & \textbf{43.45} $^{\pm 1.23}$ \\
KazNERD                             & kaz      & \textbf{41.13} $^{\pm 2.40}$  & \textbf{41.13} $^{\pm 2.40}$  \\
KIND                                & ita      & 69.22 $^{\pm 0.66}$ & \textbf{69.29} $^{\pm 0.62}$ \\
L3Cube MahaNER                      & mar      & \textbf{43.51} $^{\pm 3.38}$ & \textbf{43.51} $^{\pm 3.38}$ \\
MasakhaNER                          & amh      & \textbf{9.67} $^{\pm 0.78}$  & \textbf{9.67} $^{\pm 0.78}$  \\
MasakhaNER                          & hau      & \textbf{76.61} $^{\pm 0.52}$ & 76.53 $^{\pm 0.60}$  \\
MasakhaNER                          & ibo      & \textbf{72.46} $^{\pm 1.78}$ & \textbf{72.46} $^{\pm 1.78}$ \\
MasakhaNER                          & kin      & \textbf{62.13} $^{\pm 1.23}$ & \textbf{62.13} $^{\pm 1.23}$ \\
MasakhaNER                          & lug      & \textbf{68.28} $^{\pm 0.44}$ & 67.87 $^{\pm 0.18}$ \\
MasakhaNER                          & luo      & \textbf{58.52} $^{\pm 1.48}$ & \textbf{58.52} $^{\pm 1.48}$ \\
MasakhaNER                          & pcm      & 62.65 $^{\pm 1.94}$ & \textbf{64.11} $^{\pm 1.53}$ \\
MasakhaNER                          & swa      & \textbf{76.13} $^{\pm 1.21}$ & \textbf{76.13} $^{\pm 1.21}$ \\
MasakhaNER                          & wol      & \textbf{63.35} $^{\pm 0.54}$ & \textbf{63.35} $^{\pm 0.54}$ \\
MasakhaNER                          & yor      & \textbf{70.32} $^{\pm 1.11}$ & \textbf{70.32} $^{\pm 1.11}$ \\
MasakhaNER 2.0                      & bam      & \textbf{59.08} $^{\pm 2.21}$ & \textbf{59.08} $^{\pm 2.21}$ \\
MasakhaNER 2.0                      & bbj      & \textbf{52.15} $^{\pm 1.25}$ & \textbf{52.15} $^{\pm 1.25}$ \\
MasakhaNER 2.0                      & ewe      & \textbf{78.02} $^{\pm 1.15}$ & \textbf{78.02} $^{\pm 1.15}$ \\
MasakhaNER 2.0                      & fon      & \textbf{69.55} $^{\pm 1.84}$ & \textbf{69.55} $^{\pm 1.84}$ \\
MasakhaNER 2.0                      & hau      & \textbf{59.32} $^{\pm 1.47}$ & \textbf{59.32} $^{\pm 1.47}$ \\
MasakhaNER 2.0                      & ibo      & \textbf{58.07} $^{\pm 3.61}$ & \textbf{58.07} $^{\pm 3.61}$ \\
MasakhaNER 2.0                      & kin      & \textbf{59.59} $^{\pm 0.30}$  & \textbf{59.59} $^{\pm 0.30}$  \\
MasakhaNER 2.0                      & lug      & \textbf{76.31} $^{\pm 0.78}$ & \textbf{76.31} $^{\pm 0.78}$ \\
MasakhaNER 2.0                      & luo      & \textbf{61.82} $^{\pm 1.39}$ & \textbf{61.82} $^{\pm 1.39}$ \\
MasakhaNER 2.0                      & mos      & \textbf{65.61} $^{\pm 0.76}$ & \textbf{65.61} $^{\pm 0.76}$ \\
MasakhaNER 2.0                      & nya      & \textbf{66.76} $^{\pm 2.46}$ & \textbf{66.76} $^{\pm 2.46}$ \\
MasakhaNER 2.0                      & pcm      & \textbf{65.51} $^{\pm 0.74}$ & \textbf{65.51} $^{\pm 0.74}$ \\
MasakhaNER 2.0                      & sna      & \textbf{50.21} $^{\pm 2.37}$ & \textbf{50.21} $^{\pm 2.37}$ \\
MasakhaNER 2.0                      & swa      & \textbf{80.07} $^{\pm 1.25}$ & \textbf{80.07} $^{\pm 1.25}$ \\
MasakhaNER 2.0                      & tsn      & \textbf{67.80} $^{\pm 3.45}$  & \textbf{67.80} $^{\pm 3.45}$  \\
MasakhaNER 2.0                      & twi      & \textbf{58.51} $^{\pm 1.07}$ & \textbf{58.51} $^{\pm 1.07}$ \\
MasakhaNER 2.0                      & wol      & \textbf{68.62} $^{\pm 0.34}$ & \textbf{68.62} $^{\pm 0.34}$ \\
MasakhaNER 2.0                      & xho      & \textbf{49.82} $^{\pm 4.46}$ & \textbf{49.82} $^{\pm 4.46}$ \\
MasakhaNER 2.0                      & yor      & \textbf{28.73} $^{\pm 0.18}$ & \textbf{28.73} $^{\pm 0.18}$ \\
MasakhaNER 2.0                      & zul      & \textbf{47.37} $^{\pm 1.49}$ & \textbf{47.37} $^{\pm 1.49}$ \\
NEMO SPMRL                          & heb      & \textbf{48.49} $^{\pm 7.21}$ & \textbf{48.49} $^{\pm 7.21}$ \\
NEMO UD                             & heb      & \textbf{44.67} $^{\pm 0.89}$ & \textbf{44.67} $^{\pm 0.89}$ \\
NorNE                               & nno      & \textbf{65.32} $^{\pm 0.55}$ & \textbf{65.32} $^{\pm 0.55}$ \\
NorNE                               & nob      & \textbf{60.78} $^{\pm 0.98}$ & \textbf{60.78} $^{\pm 0.98}$ \\
RONEC                               & ron      & \textbf{40.90} $^{\pm 0.21}$  & 40.35 $^{\pm 0.76}$ \\
SLI Galician Corpora                & glg      & \textbf{64.53} $^{\pm 0.52}$ & \textbf{64.53} $^{\pm 0.52}$ \\
ssj500k                             & slv      & \textbf{49.36} $^{\pm 1.46}$ & 48.43 $^{\pm 1.66}$ \\
ThaiNNER                            & tha      & \textbf{32.10} $^{\pm 1.04}$  & \textbf{32.10} $^{\pm 1.04}$  \\
TurkuNLP                            & fin      & \textbf{65.82} $^{\pm 0.79}$ & \textbf{65.82} $^{\pm 0.79}$ \\
Tweebank                            & eng      & \textbf{61.39} $^{\pm 0.94}$ & \textbf{61.39} $^{\pm 0.94}$ \\
UNER Chinese GSD                    & cmn      & \textbf{63.29} $^{\pm 1.31}$ & \textbf{63.29} $^{\pm 1.31}$ \\
UNER Chinese GSDSIMP                & cmn      & \textbf{65.91} $^{\pm 1.27}$ & \textbf{65.91} $^{\pm 1.27}$ \\
UNER English EWT                    & eng      & \textbf{69.44} $^{\pm 1.70}$  & 69.26 $^{\pm 1.54}$ \\
UNER Maghrebi Arabic & arq      & \textbf{50.66} $^{\pm 4.64}$ & \textbf{50.66} $^{\pm 4.64}$ \\
UNER Portuguese-Bosque              & por      & \textbf{76.63} $^{\pm 0.27}$ & \textbf{76.63} $^{\pm 0.27}$ \\
UNER Slovak SNK                     & slk      & \textbf{67.30} $^{\pm 0.76}$  & \textbf{67.30} $^{\pm 0.76}$  \\
UNER Swedish Talkbanken             & swe      & \textbf{56.92} $^{\pm 2.09}$ & \textbf{56.92} $^{\pm 2.09}$ \\
WikiGoldSK                          & slk      & \textbf{71.22} $^{\pm 0.91}$ & 71.02 $^{\pm 1.06}$ \\ 
WNUT17                              & eng      &  \textbf{53.97} $^{\pm 0.45}$  & 52.28 $^{\pm 0.95}$ \\
\midrule
Mean & & 60.55$^{\pm 1.64}$ & 60.66$^{\pm 1.63}$ \\
\bottomrule
\end{tabular}
}
\centering
\caption{Results for 5-shot demonstrations with LLMs. Each configuration was run with three random seeds.}
\label{tab:llm-results}
\end{table*}

\begin{table*}[tbh]
\footnotesize
\centering
\resizebox{\textwidth}{!}{
\begin{tabular}{lllllrrlll}
\toprule
Language & Code  & Family & Branch & \begin{tabular}[c]{@{}l@{}}Script \\ (in Data)\end{tabular} & \multicolumn{1}{l}{\begin{tabular}[c]{@{}l@{}}Spkrs.\\ ($10^6$)\end{tabular}} & \multicolumn{1}{l}{\begin{tabular}[c]{@{}l@{}}Wikipedia \\ Articles\end{tabular}} & \begin{tabular}[c]{@{}l@{}}XLM-R\\ Train\end{tabular} & \begin{tabular}[c]{@{}l@{}}mBERT \\ Train\end{tabular} & \begin{tabular}[c]{@{}l@{}}Glot500 \\ Train\end{tabular} \\
\midrule
Amharic & amh & Indo-European & Semitic & Ge'ez & 35 & 15,370 & \checkmark & &\checkmark \\
Arabic & ara & Afro-Asiatic & Semitic & Arabic & 380 & 1,242,904 & \checkmark & \checkmark &\checkmark \\
Bambara & bam & Niger-Congo & Mande & Latin & 4.2 & 840 && &\checkmark\\
Bavarian German & bar & Indo-European & Germanic & Latin & 15 & 27,169 && \checkmark &\checkmark\\
Ghomálá’ & bbj & Niger-Congo & Bantoid & Latin & 0.4 & 0 && &\\
Catalan & cat & Indo-European & Romance & Latin & 4.1 & 761,156 & \checkmark & \checkmark & \checkmark \\
Mandarin Chinese & cmn & Sino-Tibetan & Sinitic & Chi. Trad./Simp. & 940 & 1,446,573 & \checkmark & \checkmark & \checkmark\\
Danish & dan & Indo-European & Germanic & Latin & 6 & 302,658 & \checkmark & \checkmark & \checkmark\\
German & deu & Indo-European & Germanic & Latin & 95 & 2,950,458 & \checkmark & \checkmark & \checkmark\\
Greek & ell & Indo-European & Hellenic & Greek & 13.5 & 240,894 & \checkmark & \checkmark & \checkmark\\
English & eng & Indo-European & Germanic & Latin & 380 & 6,895,998 & \checkmark & \checkmark & \checkmark\\
Basque & eus & Isolate & Isolate & Latin & 0.8 & 445,654 & \checkmark & \checkmark & \checkmark\\
Éwé & ewe & Niger-Congo & Volta-Niger & Latin & 5 & 951 &&& \checkmark\\
Persian Farsi & fas & Indo-European & Indo-Iranian & Perso-Arabic & 91 & 1,038,033 &\checkmark&\checkmark & \checkmark\\
Finnish & fin & Uralic & Finnic & Latin & 5 & 581,741 & \checkmark & \checkmark & \checkmark\\
Fon & fon & Niger-Congo & Volta-Niger & Latin & 2.3 & 2,059 && & \checkmark\\
Galician & glg & Indo-European & Romance & Latin & 2.4 & 214,945 & \checkmark & \checkmark & \checkmark\\
Hausa & hau & Afro-Asiatic & Chadic & Latin & 54 & 50,534 & \checkmark & & \checkmark\\
Hebrew & heb & Afro-Asiatic & Semitic & Hebrew & 5 & 363,721 & \checkmark & \checkmark & \checkmark\\
Hindi & hin & Indo-European & Indo-Aryan & Devanagari & 345 & 163,371 & \checkmark & \checkmark & \checkmark\\
Croatian & hrv & Indo-European & Slavic & Latin & 5.1 & 222,728 & \checkmark & \checkmark & \checkmark\\
Igbo & ibo & Niger-Congo & Volta-Niger & Latin & 31 & 36,914 && & \checkmark\\
Italian & ita & Indo-European & Romance & Latin & 65 & 1,886,223 & \checkmark & \checkmark & \checkmark\\
Japanese & jpn & Japonic & Japanese & Kana/Kanji & 123 & 1,433,365 & \checkmark & \checkmark & \checkmark \\
Kazakh & kaz & Turkic & Kipchak & Cyrillic & 16.7 & 237,780 & \checkmark & \checkmark & \checkmark\\
Kinyarwanda & kin & Niger-Congo & Bantu & Latin & 15 & 7,821 && & \checkmark\\
Luganda & lug & Niger-Congo & Bantu & Latin & 5.6 & 3,337 && & \checkmark\\
Luo & luo & Nilo-Saharan & Nilotic & Latin & 4.2 & 0 &&& \checkmark\\
Marathi & mar & Indo-European & Indo-Aryan & Devanagari & 83 & 98,164 & \checkmark & \checkmark & \checkmark\\
Mossi & mos & Niger-Congo & Gur & Latin & 6.5 & 0 &&& \checkmark\\
Nepali & nep & Indo-European & Indo-Aryan & Devanagari & 19 & 31,357 & \checkmark & \checkmark & \checkmark\\
Dutch & nld & Indo-European & Germanic & Latin & 25 & 2,169,462 & \checkmark & \checkmark & \checkmark\\
Norwegian (Nynorsk) & nno & Indo-European & Germanic & Latin & 4.3 & 171,312 & \checkmark & \checkmark & \checkmark\\
Norwegian (Bokmål) & nob & Indo-European & Germanic & Latin & 4.3 & 636,583 & \checkmark & \checkmark & \checkmark\\
Chichewa & nya & Niger-Congo & Bantu & Latin & 7 & 1,035 && & \checkmark\\
Naija & pcm & English Creole & English Creole & Latin & 4.7 & 1,243 &&& \checkmark\\
Portuguese & por & Indo-European & Romance & Latin & 260 & 1,134,982 & \checkmark & \checkmark & \checkmark \\
Algerian Arabic & arq & Afro-Asiatic & Semitic & Latin & 88 & 0 && &\\
Romanian & ron & Indo-European & Romance & Latin & 25 & 493,880 & \checkmark & \checkmark & \checkmark\\
Slovak & slk & Indo-European & Slavic & Latin & 5 & 250,676 & \checkmark & \checkmark & \checkmark\\
Slovenian & slv & Indo-European & Slavic & Latin & 2.5 & 187,001 & \checkmark & \checkmark & \checkmark\\
chiShona & sna & Niger-Congo & Bantu & Latin & 6.5 & 11,448 && & \checkmark\\
Spanish & spa & Indo-European & Romance & Latin & 500 & 1,983,918 & \checkmark & \checkmark & \checkmark\\
Kiswahili & swa & Niger-Congo & Bantu & Latin & 5.3 & 84,161 & \checkmark & \checkmark & \checkmark\\
Swedish & swe & Indo-European & Germanic & Latin & 10 & 2,596,219 & \checkmark & \checkmark & \checkmark\\
Thai & tha & Kra-Dai & Tai & Thai & 21 & 167,460 & \checkmark & & \checkmark\\
Setswana & tsn & Niger-Congo & Bantu & Latin & 5.2 & 1,889 && & \checkmark\\
Akan/Twi & twi & Niger-Congo & Kwa & Latin & 8.9 & 0 && & \checkmark\\
Wolof & wol & Niger-Congo & Senegambian & Latin & 7.1 & 1,704 &&& \checkmark\\
isiXhosa & xho & Niger-Congo & Bantu & Latin & 8 & 2,107 & \checkmark & & \checkmark\\
Yoruba & yor & Niger-Congo & Volta-Niger & Latin & 45 & 34,397 && \checkmark & \checkmark\\
Zulu & zul & Niger-Congo & Bantu & Latin & 13 & 11,539 && & \checkmark\\
\bottomrule
\end{tabular}
 }
\caption{Language information for the included datasets.}
\label{tab:langs}
\end{table*}

\begin{table}[tb]
\centering
\footnotesize
\begin{tabular}{lr}
\toprule
Entity Type & Count \\
\midrule
ADAGE & 197 \\
ART & 6,547 \\
ART-DERIV & 2 \\
ART-PART & 9 \\
CARDINAL & 38,290 \\
CONTACT & 202 \\
CORPORATION & 321 \\
CREATIVE\_WORK & 387 \\
DATE & 104,510 \\
DATETIME & 9,614 \\
DERIV & 1,176 \\
DESIGNATION & 980 \\
DISEASE & 1,273 \\
EVENT & 7,785 \\  
EVENT-DERIV & 6 \\
EVENT-PART & 9 \\
FACILITY & 8,825 \\ 
FESTIVAL & 266 \\
GAME & 1,762 \\
GPE & 41,915 \\
GPE-LOC & 5,104 \\
GPE-ORG & 938 \\
GROUP & 468 \\
LANG-DERIV & 64 \\
LANG-PART & 6 \\
LANGUAGE & 7,127 \\
LAW & 857 \\
LITERATURE & 847 \\
LOC & 412,987 \\ 
LOC-DERIV & 3,871 \\
LOC-PART & 699 \\
MEASURE & 6,752 \\
MISC & 40,901 \\
MISC-DERIV & 292 \\
MISC-PART & 253 \\
MONEY & 7,371 \\
MOVEMENT & 65 \\
NON\_HUMAN & 8 \\
NORP & 15,495 \\
NUM & 57,371 \\
ORDINAL & 8,061 \\
ORG & 245,855 \\ 
ORG-DERIV & 85 \\
ORG-PART & 1,101 \\
PER & 329,843 \\ 
PER-DERIV & 614 \\
PER-PART & 251 \\
PERCENT & 1,907 \\
PERCENTAGE & 4,284 \\
PERIOD & 1,188 \\
PET\_NAME & 18 \\
PHONE & 2 \\
POSITION & 6,142 \\
PRODUCT & 6,133 \\ 
PROJECT & 2,111 \\
QUANTITY & 7,496 \\
RELIGION & 1,168 \\
RELIGION-DERIV & 5 \\
TIME & 22,874 \\
TITLE\_AFFIX & 322 \\
\midrule
Total & 2,816,304 \\
\bottomrule
\end{tabular}
\caption{Counts of names of each entity type in the standardized version of OpenNER.}
\label{tab:entity_type_counts}
\end{table}

\begin{table*}[tbh]
\centering
\footnotesize
\begin{tabular}{@{}ll@{}}
\toprule
Before Mapping                             & After Mapping \\ \midrule
PER, Person, PERSON, per, NEP, pers, person                           & PER                        \\
ORG, Organization, ORGANIZATION, ORGANISATION, org, NEO & ORG                        \\
LOC, Location, LOCATION, loc, NEL, location                       & LOC                        \\
MISC, MIS, OTH, MISCELLANEOUS, misc                     & MISC                       \\
DATE, Date, NED                                         & DATE                       \\
TIMEX, TIME, NETI                                       & TIME                       \\
DATETIME                                                & DATETIME                   \\
EVENT, Event, EVT, EVE, event                                  & EVENT                      \\
PERIOD                                                  & PERIOD                     \\
NUMEX, NUMERIC, NUM                                     & NUM                        \\
CARDINAL                                                & CARDINAL                   \\
ORDINAL                                                 & ORDINAL                    \\
PERCENTAGE                                              & PERCENTAGE                 \\
GPE                                                     & GPE                        \\
GPE\_LOC                                                & GPE-LOC                    \\
GPE\_ORG                                                & GPE-ORG                    \\
FACILITY, FAC, fac                                           & FACILITY                   \\
LANGUAGE, LANG, ANG                                          & LANGUAGE                   \\
MONEY                                                   & MONEY                      \\
NORP, NAT\_REL\_POL                                     & NORP                       \\
PRODUCT, PROD, PRO, DUC, pro, product                                 & PRODUCT                    \\
QUANTITY                                                & QUANTITY                   \\
PROJECT                                                 & PROJECT                    \\
ART, WORK\_OF\_ART, WOA                                 & ART                        \\
CONTACT                                                 & CONTACT                    \\
DISEASE                                                 & DISEASE                    \\
FESTIVAL                                                & FESTIVAL                   \\
GAME                                                    & GAME                       \\
LAW                                                     & LAW                        \\
LITERATURE                                              & LITERATURE                 \\
NON\_HUMAN                                              & NON\_HUMAN                 \\
RELIGION                                                & RELIGION                   \\
ADAGE                                                   & ADAGE                      \\
POSITION                                                & POSITION                   \\
PERderiv, deriv-per                                     & PER-DERIV                  \\
ORGderiv                                                & ORG-DERIV                  \\
LOCderiv                                                & LOC-DERIV                  \\
OTHderiv, MISCderiv                                     & MISC-DERIV                 \\
PERpart                                                 & PER-PART                   \\
ORGpart                                                 & ORG-PART                   \\
LOCpart                                                 & LOC-PART                   \\
OTHpart, MISCpart                                       & MISC-PART                  \\
EVENTderiv                                              & EVENT-DERIV                \\
EVENTpart                                               & EVENT-PART                 \\
LANGderiv                                               & LANG-DERIV                 \\
LANGpart                                                & LANG-PART                  \\
WOAderiv                                                & ART-DERIV                  \\
WOApart                                                 & ART-PART                   \\
RELIGIONderiv                                           & RELIGION-DERIV             \\
DRV                                                     & DERIV                      \\
ED                                                      & DESIGNATION                \\
NEM                                                     & MEASURE                    \\ 
group & GROUP \\
corporation & CORPORATION \\
\bottomrule
\end{tabular}
\caption{Mapping from original entity types to standardized entity types.}
\label{tab:map-unify-types}
\end{table*}

\subsection{Hyperparameters and Computational Resources}
\label{app:hyperparams}
Models are trained using HuggingFace's transformers package \citep{wolf-etal-2020-transformers}. 
We use an encoder model with a TokenClassification head (linear layer). 
No CRF is used in our experiments.
The first subtoken of each word is used for the label. 
Hyperparameters for fine-tuning were set to a learning rate of 5.0e-5, 10 epochs of fine-tuning, weight decay of 0.05, a batch size of 16, and a warm-up ratio of 0.1.
Model sizes in terms of number of parameters are as follows:
XLM-R: 279 million, 
mBERT: 179 million, 
Glot500: 395 million.

Experiments were run on a SLURM cluster with 32 NVIDIA GPUs (16 RTX A5000, 8 A40, 8 RTX 6000 Ada Generation). Approximately 400 GPU hours were used to train and evaluate the mBERT, XLM-R, and Glot500 models, and approximately 384 GPU hours were used to evaluate the LLM models.

\subsection{LLM Few-Shot Prompt Example}
\label{app:llm}

We conducted a small set of trials with Aya-Expanse 32b and CoNLL-03 English to determine which approach to prompting the LLMs appeared to perform best. 

We explored prompting the model to output each token and its BIO tag, but this performed poorly and it was challenging to map the output back to the original text. 

We also experimented with prompting the model to return a JSON object with the three entity types and lists of entities.
For example, we expect an output from the LLM such as the following:

\begin{verbatim}
{
  "Person": [],
  "Location": ["AL-AIN",
               "United Arab Emirates"],
  "Organization": []
}
\end{verbatim}

Using JSON as the response had an overall F1 of 61.46. 

We then prompted the model to return the original sentence ``inline'' with special markers for the start and end of the entity text.
This inline approach scored 67.42, notably better than the response as JSON.
Since the inline approach worked best, we further attempted this with 5-shot demonstrations. 
 Inline with demonstrations was our best performing approach at 74.86 F1, so we continued with this approach for our experiments with LLMs across all datasets in OpenNER.

Demonstrations were randomly selected from the training dataset on a per-language per-dataset basis.
Example sentences were filtered by length to include only examples with more than 8 tokens.
Examples were chosen to show one demonstration with no named entities and the rest of the examples were randomly selected but required to have at least one named entity.

Hyperparameters for LLMs were chosen from the generation configurations from the models themselves.
Temperature was set at 0.3 following the generation configuration in Aya-Expanse.

To conduct 5-shot demonstrations, we provided the examples as demonstrations in a conversation template. 
A conversation history with 5 turns between the user and the chatbot demonstrates how the chatbot should respond to the provided input.
Then the user input is provided with the start token for the chatbot to complete its response. 
An example of the conversation with examples is shown in Table \ref{tab:llm-prompt}. 
The target sentence to be labeled is the final example ``It is a place in Argentina lol''.

\section{Datasets not Included Due to Quality Concerns}
\label{app:not-included-data}

\citet{singh-2008-named} created a dataset for Southeast Asian languages available at \url{http://ltrc.iiit.ac.in/ner-ssea-08/index.cgi?topic=5}.
Named entity tags should all be of the form \texttt{<ne=X>} where X is one of the types in the ontology. However, many of the types do not match the ontology, and some entity types are blank. For example, correct annotations might be of the form \texttt{<ne=NEP>}, \texttt{<ne=NEL>}, and \texttt{<ne=NEM>}, but there are examples such as \texttt{<ne=k1>} and \texttt{<ne=>}.

\citet{bahad-etal-2024-fine} created an NER dataset with 4 Indic languages. 
The dataset is available at \url{https://github.com/ltrc/IL-NER/tree/main/Datasets} and \url{https://huggingface.co/datasets/Sankalp-Bahad/NER-Dataset}.
We were unable to include this dataset due to errors in the formatting of the labels.
Each named entity in BIO or BIOES encoding must have a state indicating whether it is Begin, Inside, Single, etc. but there are 1,662 examples of tags where the state is missing such as \texttt{-NEO} or \texttt{-NEL} across all four languages. 
It is nontrivial to correct these errors, since adjacent entities may need different states and corrections cannot be made automatically without having to examine each example.
For Hindi, the test set has 30 invalid labels. The training data has at least one missing token and 150 invalid labels. The dev set has 16 invalid labels.
For Odia, the training set has 28 invalid labels, the dev set has 1 invalid label, and test set has 1 invalid label. 
For Telugu, the train set has 1,073 invalid labels, the dev has 181 invalid labels, and the  test  has 166 invalid labels.
For Urdu, the train set has 12 invalid labels, the dev set has 3 invalid labels, and the test set has 1 invalid label.

The myNER dataset \citep{thant2025mynercontextualizedburmesenamed} consists of NER data in Burmese. 
myNER uses the BIOES tagging scheme and has validation errors, the majority of which appear to occur around parentheses and hyphens as part of dates. Unfortunately, these invalid transitions are too many and not consistent enough to fix automatically.
The train set contains 2,388 invalid transitions, the dev set 312, and the test set 280. 

\citet{karim2019step} created a dataset for NER in Bangla. 
Unfortunately, this dataset has 311 invalid label transitions.  
While it is sometimes possible to repair invalid transitions automatically (for example, \texttt{O} to \texttt{I} can be automatically converted to \texttt{O} to \texttt{B}), the training dataset has 26 transitions which have invalid transitions with differing adjacent types. 
The dev set has 3 of these invalid transitions with adjacent types, and the test set has 4. 
These invalid transitions can require knowledge of the language in order to ensure that the correct interpretation is chosen.

\section{Ontologies}
\label{app:ontologies}

\subsection{CoNLL-Derived Ontologies}

The CoNLL-02 corpus \citep{tjong-kim-sang-2002-introduction} consists of Spanish and Dutch newswire data and introduces the LOC/ORG/PER/MISC tagset adapted by many other corpora in this collection.

The AnCora corpus \citep{taule-etal-2008-ancora} performed multi-level linguistic annotation on news data in both Catalan and Spanish. The NER component of the corpus uses the CoNLL-02 ontology, but with OTHER instead of MISC, and with the addition of NUMBER and DATE.

The AQMAR corpus \citep{mohit-etal-2012-recall} contains NER data sourced from Wikipedia articles in Arabic.
We use a version\footnote{\url{https://github.com/LiyuanLucasLiu/ArabicNER/tree/master}} with fixes of invalid label sequences by \citet{liu-etal-2019-arabic}. 

The BarNER corpus \citep{peng-etal-2024-sebastian} consists of NER annotation on Bavarian Wikipedia and Twitter data, using CoNLL core types in addition to LANG (language), RELIGION, EVENT, and WOA (work of art). Each entity type can also appear with suffix -part or -deriv, to capture the nominal derivation and compounding common in the language.

The DaNE corpus \citep{hvingelby-etal-2020-dane} is named entity annotation as an extension of Universal Dependencies. 
The underlying data is the PAROLE corpus \citep{keson1998vejledning}, which was built from paragraphs from a Danish Dictionary.

EIEC \citep{alegria2006lessons}\footnote{\url{http://www.ixa.eus/node/4486?language=en}} is a corpus of Basque newswire.

EverestNER is an NER corpus of news articles \citep{niraula2022named}. 
It uses the CoNLL-02 ontology without MISC but with EVENT and DATE types.

The GermEval2014 corpus \citep{benikova2014germeval} contains data from the 2014 GermEval NER shared task which includes newswire and German Wikipedia data.
The tagset used to annotate this corpus is very similar to the CoNLL-02 one, however the MISC type is renamed OTH (other) and subtypes are introduced.
These subtypes occur in the form of TYPEderiv and TYPEpart, with deriv signifying a derivation of the original type and part a named entity that is part of a larger entity.

HiNER \citep{murthy-etal-2022-hiner} is a Hindi dataset that is made up of newswire and data from the tourism domain.
The tagset used corpus is based on the CoNLL-02 ontology, with additional custom tags added to further specify categories encompassed by the MISC type (FESTIVAL, GAME, LANGUAGE, LITERATURE, RELIGION).

The KIND corpus \citep{paccosi-palmero-aprosio-2022-kind} is a multi-domain Italian corpus which uses the CoNLL-02 types without MISC. The domains included are literature, political discourse, and Wikinews. During preprocessing the train and test sets across all domains were concatenated.
The dataset did not contain a development set.

hr500k is corpus of morpho-syntactic annotation on Croatian web and news data \citep{ljubesic-etal-2016-new}. 
L3Cube-MahaNER \citep{litake-etal-2022-l3cube} is a Marathi news dataset for named entity recognition. 

The MasakhaNER version 1.0 dataset \citep{adelani-etal-2021-masakhaner} is a multilingual dataset that contains local news data in 10 different African languages.
It uses the CoNLL-02 types without MISC and with the addition of DATE.
We also include MasakhaNER 2.0 \citep{adelani-etal-2022-masakhaner}, which uses the same ontology but covers additional languages.

NEMO$^2$ \citep{bareket-tsarfaty-2021-neural} consists of both morpheme and token based NER annotation on the Hebrew Treebank \citep{simaan_hebrew_treebank}, based on the CoNLL-02 guidelines but also adopting GPE, facility, work of art, language, product, and event types. Parallel annotations are provided for the UD version of the Hebrew Treebank \citep{sade-etal-2018-hebrew}.

NorNE \citep{jorgensen-etal-2020-norne} is an NER corpus containing both Norwegian Bokmål (nob) and Nynorsk (nno) standards.
The corpus is mainly news data, but also contains government reports, parliamentary transcripts and blog posts.
The ontology is CoNLL-02-like but includes GPE\_LOC and GPE\_ORG. 
EVT and PROD are also included. 

A subset of the SLI Galician CTG corpora \citep{agerri-etal-2018-developing}, from the news and environmental sciences domains, has been annotated for NER, following the CoNLL guidelines.

ssj500k \citep{dobrovoljc-etal-2017-universal} uses the CoNLL-02 ontology. It contains data from fiction, non-fiction, periodical and Wikipedia texts.
Since canonical splits did not appear to exist we created splits in a 80/10/10 manner following the approach used in the GitHub repository.\footnote{\url{https://github.com/TajaKuzman/NER-recognition/blob/master/create_NER_task_files.py}}

WikiGoldSK \citep{suba-etal-2023-wikigoldsk} is Slovak NER on Wikipedia data with the CoNLL-02 ontology. 

The Turku NER corpus \citep{luoma-etal-2020-broad} is a Finnish corpus that builds on the original Universal Dependencies Finnish corpus \citep{nivre-etal-2016-universal}, which is made up of multi-domain data including news, web, legal, fiction and political data. 
It uses the CoNLL-02 tags LOC, PER and ORG, but not MISC. The types PRO (Product), DATE and EVENT are also included.

The Tweebank NER dataset \citep{jiang-etal-2022-annotating} is an English dataset developed by annotating the Tweebank V2 \citep{liu-etal-2018-parsing}, the main universal dependency treebank for English Twitter NLP tasks.
Tweebank uses standard CoNLL-02 tags. 

WNUT17 \citep{derczynski-etal-2017-results} annotates web text with emerging entities of 6 entity types which further subdivide ORG into group and corporation.

The ArmanPersoNERCorpus \citep{poostchi-etal-2016-personer} in Persian extends CoNLL ontology with facility, event, and product. 

We also included several UNER datasets: Chinese GSD \citep{ud_chinese_gsd,ud_chinese_gsdsimp},
English EWT \citep{silveira-etal-2014-gold},
Maghrebi \citep{seddah-etal-2020-building},
Portuguese Bosque \citep{rademaker-etal-2017-universal},
SNK \citep{Zeman2017SlovakDT}, and
Swedish Talkbanken \citep{mcdonald-etal-2013-universal}.

\subsection{OntoNotes-Derived Ontologies}

elNER \citep{elner-2020} performs NER annotations on Greek news data based on the OntoNotes ontology, but also provides a CoNLL-derived version by merging and filtering types.
We use the OntoNotes version of their data.

NER labels were added to Japanese-GSD-UD \citep{asahara-etal-2018-universal} by Meganon labs.\footnote{\url{https://github.com/megagonlabs/UD_Japanese-GSD}}
The ontology has 21 entity types largely following OntoNotes with the addition of TITLE\_AFFIX, MOVEMENT, PHONE, and PET\_NAME, and the corpus is made up of Wikipedia data.

The KazNERD corpus \citep{yeshpanov-etal-2022-kaznerd} uses the OntoNotes ontology for annotation of Kazakh.

RONEC \citep{dumitrescu-avram-2020-introducing} uses an OntoNotes-like ontology but with some types collapsed (i.e. DATETIME, NAT\_REL\_POL) and some missing (PROD, LAW). The data included in this dataset is collected from news texts.

Thai NNER \citep{buaphet-etal-2022-thai} uses a fine-grained NER ontology and 10 coarse-grained top-level types.
The coarse-grained types are from the OntoNotes ontology. 
Because Thai NNER is nested NER, we make use of only the top level entity span. 
The data is made up of news articles and restaurant reviews.
The dataset is syllable and document segmented, but not sentence segmented.
This segmentation is why Thai appears to have a comparatively small number of sentences.

\section{Additional Tables}
Additional tables are included on the following pages.

\begin{table*}[tb]
\centering 
\footnotesize
\begin{tabular}{p{0.90\linewidth}}
\toprule
\textless{}|START\_OF\_TURN\_TOKEN|\textgreater{}\textless{}|SYSTEM\_TOKEN|\textgreater{} \\

You are a powerful conversational AI trained by Cohere to help people. You are augmented by a number of tools, and your job is to use and consume the output of these tools to best help the user. You will see a conversation history between yourself and a user, ending with an utterance from the user. You will then see a specific instruction instructing you what kind of response to generate. When you answer the user's requests, you cite your sources in your answers, according to those instructions.  \\        
\textless{}|END\_OF\_TURN\_TOKEN|\textgreater{} \\

\textless{}|START\_OF\_TURN\_TOKEN|\textgreater{}\textless{}|USER\_TOKEN|\textgreater{} \\
Find names of persons, organizations or locations. Label the following sentence with labels where the name is enclosed with the entity type PER, ORG or LOC and @@ \#\#. For example PER @@ John Smith \#\#, or ORG @@ Springfield University \#\# or LOC @@ United Kingdom \#\# .
Find named entities in the following sentence:
Check out the kids ' space as soon as you get onboard . \\
\textless{}|END\_OF\_TURN\_TOKEN|\textgreater{} \\
\textless{}|START\_OF\_TURN\_TOKEN|\textgreater{}\textless{}|CHATBOT\_TOKEN|\textgreater{} \\
Check out the kids ' space as soon as you get onboard . \\
\textless{}|END\_OF\_TURN\_TOKEN|\textgreater{} \\

\textless{}|START\_OF\_TURN\_TOKEN|\textgreater{}\textless{}|USER\_TOKEN|\textgreater{}

Find names of persons, organizations or locations. Label the following sentence with labels where the name is enclosed with the entity type PER, ORG or LOC and @@ \#\#. For example PER @@ John Smith \#\#, or ORG @@ Springfield University \#\# or LOC @@ United Kingdom \#\# . \\
Find named entities in the following sentence: \\
A pair of military pilots embarked on China 's second manned space flight early today , hurtling into orbit on a test mission that could span five days . \\
\textless{}|END\_OF\_TURN\_TOKEN|\textgreater{} \\
\textless{}|START\_OF\_TURN\_TOKEN|\textgreater{}\textless{}|CHATBOT\_TOKEN|\textgreater{} \\
A pair of military pilots embarked on LOC @@ China \#\# 's second manned space flight early today , hurtling into orbit on a test mission that could span five days . \\
\textless{}|END\_OF\_TURN\_TOKEN|\textgreater{} \\
\textless{}|START\_OF\_TURN\_TOKEN|\textgreater{}\textless{}|USER\_TOKEN|\textgreater{} \\
Find names of persons, organizations or locations. Label the following sentence with labels where the name is enclosed with the entity type PER, ORG or LOC and @@ \#\#. For example PER @@ John Smith \#\#, or ORG @@ Springfield University \#\# or LOC @@ United Kingdom \#\# . \\
Find named entities in the following sentence: \\
Very recently , US forces in Al - Anbar made a terrible blunder by accidentally killing Abdul - Razaq Inad Al - Gu'ud , Sheikh of the Al - Bu Nimr clan from the powerful Dulaym tribe . \\
\textless{}|END\_OF\_TURN\_TOKEN|\textgreater{} \\
\textless{}|START\_OF\_TURN\_TOKEN|\textgreater{}\textless{}|CHATBOT\_TOKEN|\textgreater{}Very recently , LOC @@ US \#\# forces in LOC @@ Al - Anbar \#\# made a terrible blunder by accidentally killing PER @@ Abdul - Razaq Inad Al - Gu'ud \#\# , Sheikh of the PER @@ Al - Bu Nimr \#\# clan from the powerful Dulaym tribe. \textless{}|END\_OF\_TURN\_TOKEN|\textgreater{} \\
\textless{}|START\_OF\_TURN\_TOKEN|\textgreater{}\textless{}|USER\_TOKEN|\textgreater{}Find names of persons, organizations or locations. Label the following sentence with labels where the name is enclosed with the entity type PER, ORG or LOC and @@ \#\#. For example PER @@ John Smith \#\#, or ORG @@ Springfield University \#\# or LOC @@ United Kingdom \#\# . \\

Find named entities in the following sentence: \\
Ordinarily , George would come in around noon ; he would ordinarily leave around 5:30 or 6:00 in the evening .\textless{}|END\_OF\_TURN\_TOKEN|\textgreater{} \\
\textless{}|START\_OF\_TURN\_TOKEN|\textgreater{}\textless{}|CHATBOT\_TOKEN|\textgreater{} \\
Ordinarily , PER @@ George \#\# would come in around noon ; he would ordinarily leave around 5:30 or 6:00 in the evening. \\
\textless{}|END\_OF\_TURN\_TOKEN|\textgreater{} \\
\textless{}|START\_OF\_TURN\_TOKEN|\textgreater{}\textless{}|USER\_TOKEN|\textgreater{}Find names of persons, organizations or locations. Label the following sentence with labels where the name is enclosed with the entity type PER, ORG or LOC and @@ \#\#. For example PER @@ John Smith \#\#, or ORG @@ Springfield University \#\# or LOC @@ United Kingdom \#\# . \\
Find named entities in the following sentence: \\
I received this draft from Niagara Mohawk Marketing , Inc. for our review .\textless{}|END\_OF\_TURN\_TOKEN|\textgreater{} \\
\textless{}|START\_OF\_TURN\_TOKEN|\textgreater{}\textless{}|CHATBOT\_TOKEN|\textgreater{} \\
I received this draft from ORG @@ Niagara Mohawk Marketing , Inc. \#\# for our review. \\
\textless{}|END\_OF\_TURN\_TOKEN|\textgreater{} \\
\textless{}|START\_OF\_TURN\_TOKEN|\textgreater{}\textless{}|USER\_TOKEN|\textgreater{} \\
Find names of persons, organizations or locations. Label the following sentence with labels where the name is enclosed with the entity type PER, ORG or LOC and @@ \#\#. For example PER @@ John Smith \#\#, or ORG @@ Springfield University \#\# or LOC @@ United Kingdom \#\# . \\
Find named entities in the following sentence: \\
It is a place in Argentina lol \\
\textless{}|END\_OF\_TURN\_TOKEN|\textgreater{} \\
\textless{}|START\_OF\_TURN\_TOKEN|\textgreater{}\textless{}|CHATBOT\_TOKEN|\textgreater{} \\           
\bottomrule
\end{tabular}
\centering
\caption{Full example of inline filled prompt template with 5--shot demonstrations.}
\label{tab:llm-prompt}
\end{table*}

\end{document}